\documentclass{article}

\usepackage{microtype}
\usepackage{graphicx}
\usepackage{subcaption}
\usepackage{booktabs} 
\usepackage{tabularx}
\usepackage{hyperref}

\newcommand{\eg}{\textit{e.g.,}}

\usepackage[preprint]{icml2026}

\usepackage{amsmath}
\usepackage{amssymb}
\usepackage{mathtools}
\usepackage{amsthm}

\usepackage[capitalize,noabbrev]{cleveref}

\theoremstyle{plain}

\theoremstyle{definition}

\theoremstyle{remark}

\usepackage[textsize=tiny]{todonotes}

\usepackage{booktabs}       
\usepackage{multirow}       
\usepackage[table]{xcolor}  
\usepackage{graphicx}
\usepackage{enumitem}
\usepackage{amsmath}
\usepackage{amssymb}
\usepackage{xcolor}
\usepackage{tcolorbox}
\usepackage{fontawesome5}
\tcbuselibrary{listings, skins, breakable}
\definecolor{citecolor}{HTML}{0071bc}
\tcbset{
  simpleprompt/.style={
    breakable, enhanced,
    boxrule=0.7pt, arc=2mm,
    colframe=black!80,              
    left=6pt,right=6pt,top=6pt,bottom=6pt,
    fonttitle=\bfseries,
  }
}
\newtcolorbox{prompt}[2][]{simpleprompt,
  colback=citecolor!6!white,        
  title={#2}, #1}
  
\usepackage{gradient-text}

\definecolor{groupbg}{RGB}{235, 235, 235}
\definecolor{oursbg}{RGB}{220, 240, 255}

\definecolor{impbg}{RGB}{200, 230, 255} 
\definecolor{imptext}{RGB}{0, 55, 130}

\newcommand{\hflogo}{\raisebox{-0.25\height}{\includegraphics[height=1.45em]{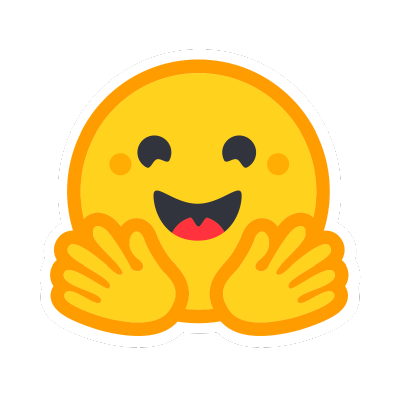}}}
\newcommand{\ghlogo}{\raisebox{-0.2\height}{\includegraphics[height=1.3em]{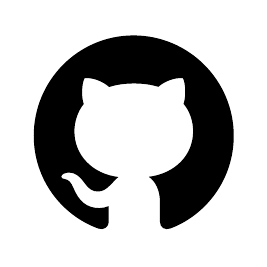}}}

\icmltitlerunning{Code2World: A GUI World Model via Renderable Code Generation}

\begin{document}
\newcommand{\kevin}[1]{\textcolor{blue}{(Kevin: #1)}}

\twocolumn[
  \icmltitle{\raisebox{-0.35\height}{
  \includegraphics[height=1cm]{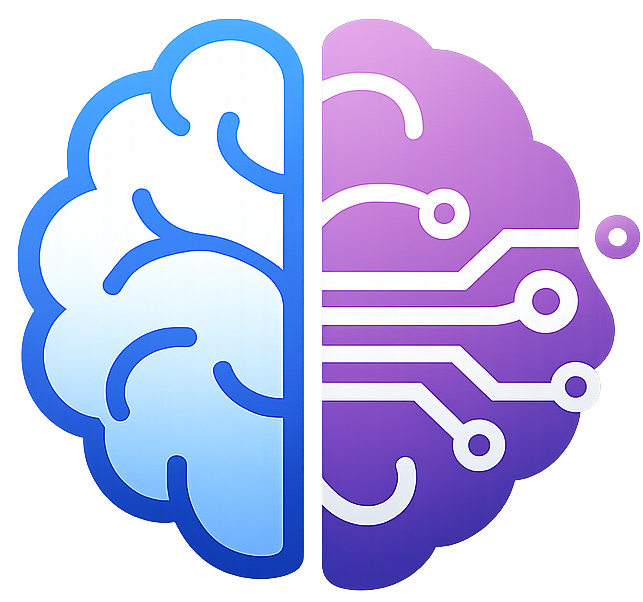}
}
Code2World: A GUI World Model via Renderable Code Generation}



  \icmlsetsymbol{equal}{*}

  \begin{icmlauthorlist}
    \icmlauthor{Yuhao Zheng}{ustc,amap,equal}
    \icmlauthor{Lian Zhong}{sysu,equal}
    \icmlauthor{Yi Wang}{amap}
    \icmlauthor{Rui Dai}{amap}
    \icmlauthor{Kaikui Liu}{amap}
    \icmlauthor{Xiangxiang Chu}{amap}
    \icmlauthor{Linyuan L\"u}{ustc}
    \icmlauthor{Philip Torr}{oxford}
    \icmlauthor{Kevin Qinghong Lin}{oxford}
    
    \vspace{1em}
    \href{https://amap-ml.github.io/Code2World}{{\color{black}\faGlobe}~\,Project Page} \quad
    \href{https://github.com/AMAP-ML/Code2World}{{\color{black}\ghlogo}~Code} \quad
    \href{https://huggingface.co/datasets/GD-ML/AndroidCode}
    {{\color{black}\faDatabase}~\,Dataset} \quad
    \href{https://huggingface.co/GD-ML/Code2World}{{\color{black}\hflogo}~Model}
  \end{icmlauthorlist}

  \icmlaffiliation{ustc}{University of Science and Technology of China}
  \icmlaffiliation{amap}{AMAP, Alibaba Group}
  \icmlaffiliation{oxford}{University of Oxford}
  \icmlaffiliation{sysu}{Sun Yat-sen University}

  \icmlcorrespondingauthor{Linyuan L\"u}{linyuan.lv@ustc.edu.cn}
  \icmlcorrespondingauthor{Kevin Qinghong Lin}{kevin.qh.lin@gmail.com}

  \icmlkeywords{Machine Learning, ICML}

  \vskip 0.3in
]



\printAffiliationsAndNotice{\icmlEqualContribution}

\begin{abstract}
Autonomous GUI agents interact with environments by perceiving interfaces and executing actions.
As a virtual sandbox, the GUI World model empowers agents with human-like foresight by enabling action-conditioned prediction.
However, existing text- and pixel-based approaches struggle to simultaneously achieve high visual fidelity and fine-grained structural controllability.
To this end, we propose \textbf{Code2World}, a vision-language coder that simulates the next visual state via \textbf{renderable code generation}. Specifically, 
to address the data scarcity problem, we construct \textbf{AndroidCode} by translating GUI trajectories into high-fidelity HTML and refining synthesized code through a visual-feedback revision mechanism, yielding a corpus of \textbf{over 80K} high-quality screen-action pairs.
To adapt existing VLMs into code prediction, we first perform SFT as a cold start for format layout following, then further apply \textbf{Render-Aware Reinforcement Learning} which uses rendered outcome as the reward signal by enforcing visual semantic fidelity and action consistency.
Extensive experiments demonstrate that Code2World-8B achieves the top-performing next UI prediction, rivaling the competitive GPT-5 and Gemini-3-Pro-Image. 
Notably, \textit{Code2World significantly enhances downstream navigation success rates in a flexible manner}, boosting Gemini-2.5-Flash by {+9.5\%} on AndroidWorld navigation.
\end{abstract}
\begin{figure}[t]
\vspace{-10pt}
  \centering
  \includegraphics[width=\columnwidth]{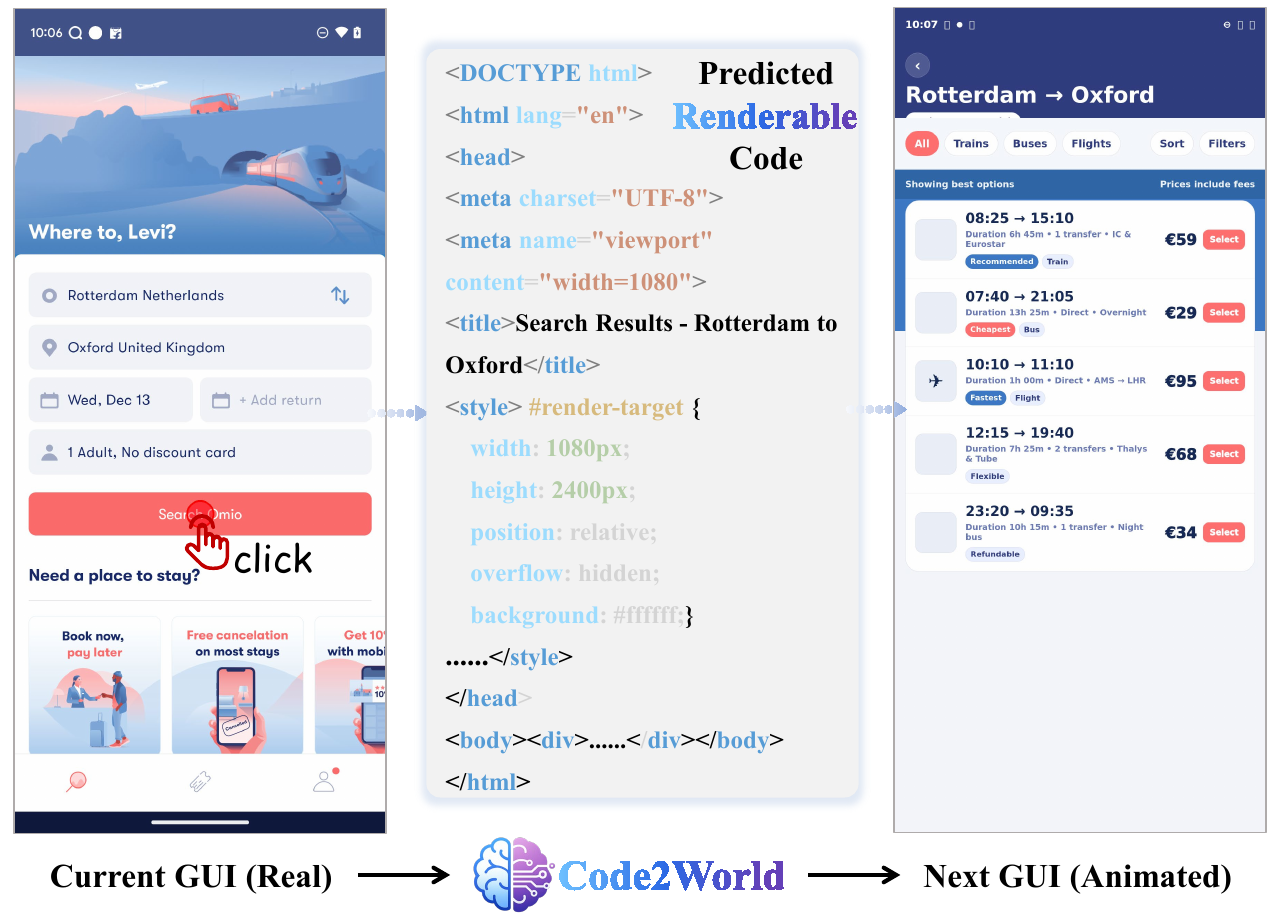}
  \vspace{-15pt}
  \caption{\textbf{Illustration of Code2World.} Given a current GUI observation and an action, Code2World predicts the next screenshot via renderable code generation.}
  \label{fig:teaser}
  \vspace{-18pt}
\end{figure}

\section{Introduction}
Recent advancements in Large Multimodal Models have revolutionized the development of autonomous GUI agents \cite{cogagent,showui,ui-tars}. These agents are designed to perceive visual interfaces and execute sequences of actions to complete complex tasks across web and mobile platforms \cite{osworld, androidcontrol}. 
Despite these promising capabilities, a significant gap with human proficiency persists in real-world scenarios, particularly for navigation tasks that demand precise reasoning and error correction \cite{wma}.

A key factor contributing to human superiority is the ability to \textit{mentally simulate the consequences of actions before execution} to evaluate potential outcomes and adjust their strategies accordingly. 
In contrast, most existing GUI agents operate actions directly without such foresight. 
Lacking this predictive capability, an erroneous execution at the current step often necessitates costly multi-step corrections or leads to immediate task failure~\cite{gui-reflection,videogui}.
This is particularly dangerous in high-risk scenarios, such as confirming payments or deleting critical data, where actions are irreversible. 
To mitigate these risks, it is essential to equip agents with a virtual sandbox to simulate the action-conditioned observation, often referred to as a \textit{world model} which has already demonstrated remarkable success in domains like embodied AI~\cite{gwm}.

Recent works have attempted to explore GUI world models under two different representation.
\textbf{Text-based approaches}~\citep{wma, webdreamer} leverage Large Language Models (LLMs) to predict state transitions via abstract natural language descriptions. 
Although these methods capture semantic intent, they fundamentally lack visual information. 
Visual feedback provides a more intuitive representation of the environment and has been shown to significantly enhance the reasoning capabilities of agents \cite{think_with_image}. 
Conversely, \textbf{Pixel-based approaches}~\cite{ui-diffuser,vimo} predominantly utilize diffusion models to synthesize future screenshots. 
However, it is non-trivial to model the precise and discrete state transitions of GUIs within a continuous pixel space.
Such approaches also face challenges in maintaining fine-grained structural controllability, particularly in text-rich interface \cite{artist}.
Ultimately, neither text nor pixel representations can simultaneously achieve high-fidelity visual simulation and precise structural controllability.

Motivated by the fact that GUI is natively created by \textbf{programmatic code} (\eg~HTML), can we explore it as a representation for learning a GUI world model?
Unlike abstract text or raw pixels, symbolic code inherently guarantees structural controllability while enabling high-fidelity visualization via deterministic rendering.
Leveraging this insight, we propose a paradigm shift by treating GUI simulation as \textit{renderable code generation}. Under this paradigm, we introduce \textbf{Code2World}, as illustrated in Figure \ref{fig:teaser}, a vision-language coder designed to predict dynamic transitions by synthesizing structured \textit{HTML code} and rendering it into the next visual state. 
However, realizing this paradigm presents three significant challenges: 
\textit{{\textbf{(i)} {Data scarcity}:}} how to curate large-scale, high-quality data that ground diverse GUI states in faithful structured code;
\textit{{\textbf{(ii)} {Screenshot-code alignment}:}} how to build a backbone and devise a learning strategy that effectively align textual code generation with rendered visual reality;
and \textit{{\textbf{(iii)} {Evaluation \& Application}:}} 
how to rigorously benchmark next UI prediction and effectively assist downstream GUI agents.

To overcome these challenges, 
\textit{{\textbf{(i)}}} we first construct \textbf{AndroidCode}, a large-scale corpus comprising over \textbf{80K} samples. Derived from the image-based GUI trajectories in AndroidControl~\cite{androidcontrol}, we synthesize their corresponding HTML representations via GPT-5. Crucially, to ensure data quality, we implement a visual-feedback revision mechanism that refines the generated code based on visual alignment checks. 
\textit{{\textbf{(ii)}}} Building on a Vision-Language Model (VLM), we begin with Supervised Fine-Tuning (SFT) as a cold start for format following, and then 
we further employ a \textbf{Render-Aware Reinforcement Learning} strategy to enhance the generalization. 
This strategy integrates dual outcome rewards to align the model with visual reality and action consistency.
\textit{{\textbf{(iii)}}} 
Finally, we establish a holistic evaluation protocol utilizing a VLM-as-a-Judge framework to benchmark functional logic and visual quality.
We further validate Code2World’s practical value by integrating it as a \textbf{plug-and-play} simulator to enhance diverse downstream GUI agents.
Extensive experiments demonstrate that Code2World-8B achieves the top-performing next UI prediction, rivaling GPT-5 and Gemini-3-Pro-Image, and significantly enhances downstream navigation success rates, boosting Gemini-2.5-Flash by \textbf{+9.5\%} on AndroidWorld \cite{androidworld}.
Our contributions are as follows.

\begin{itemize}[leftmargin=*, topsep=0pt, partopsep=0pt, itemsep=2pt, parsep=0pt]

    \item We propose Code2World, a novel VLM-based GUI World Model that predicts dynamic transitions via renderable code generation. By utilizing structured HTML as the native representation, our framework simultaneously achieves high-fidelity visual simulation and precise structural controllability.

    \item We present AndroidCode, a high-fidelity corpus comprising over 80K samples. We employ a synthesis pipeline with visual-feedback revision to ensure high data quality.
   
    \item We introduce a novel Render-Aware Reinforcement Learning strategy to align the textual output with visual reality and ensuring the generalization. 

    \item Extensive experiments demonstrate that Code2World achieves top performance in next GUI prediction. As a plug-and-play simulator, it significantly enhances the capabilities of downstream GUI agents, yielding consistent improvements in both offline and online navigation tasks.

\end{itemize}
\section{Related Work}

\subsection{GUI Agents}
The advent of Large Multimodal Models (LMMs) has fundamentally shifted GUI agents from metadata dependency to direct pixel-level perception, establishing robust baselines for generalist control~\cite{cogagent,showui,ui-tars}. 
To tackle the complexity of long-horizon navigation, recent research has augumented these agents with advanced deliberative capabilities. Innovations in advanced grounding~\cite{uivision,groundcua}, efficient context management~\cite{mga, gui-rise} and self-reflection mechanisms~\cite{gui-reflection, ui-genie} enable agents to maintain goal consistency and autonomously refine strategies across extended trajectories. 
Currently, the rapid evolution of Reinforcement Learning (RL) has empowered LMMs with advanced reasoning and planning capabilities in various domains~\cite{xiong2025hs,harderisbetter,yuan2025video,wang2026urban, treesearch, gpg}.
In the context of GUI automation, RL has been widely adopted to optimize decision logic, ranging from reasoning-enhanced ``R1-style'' fine-tuning~\cite{gui-r1, ui-r1} to online policy optimization~\cite{mobilegui-rl, arpo}.
However, these approaches optimize the agent's policy in the actual environment. Code2World shifts the focus from the agent to the environment, functioning as a learnable virtual sandbox. Unlike RL for the agent itself, we employ a render-aware RL algorithm to optimize our simulator.

\subsection{GUI Environments and World Models}
Autonomous GUI agents operate within interactive environments, executing actions to elicit feedbacks~\cite{androidworld, osworld}. 
World models internalize these dynamics to function as efficient virtual sandboxes for agent planning. Existing research typically adopts two output modalities.
\textit{(i) Text-based approaches} leverage Large Language Models (LLMs) to predict state transitions via symbolic abstraction, ranging from natural language descriptions of state differentials~\cite{wma, mobileworldbench,aui} and structural DOM updates~\cite{webdreamer, ui-simulator} to textual sketches preserving spatial layouts~\cite{mobiledreamer}.
Conversely, \textit{(ii) Pixel-based approaches} utilize image generation models to synthesize future screenshots directly, often adapting diffusion architectures to generate UI layouts and frame sequences~\cite{ui-diffuser, neuralos}, or decoupling text and layout generation~\cite{vimo} for enhanced fidelity.
However, text-based methods discard critical spatial-visual details, while pixel-based synthesis struggles with structural controllability and text-rich scenarios. To bridge this gap, we propose a \textit{(iii) renderable code approach} that ensures both visual fidelity and structural integrity.

\begin{figure*}[t]
\vspace{-8pt}
  \centering
  \includegraphics[width=\textwidth]{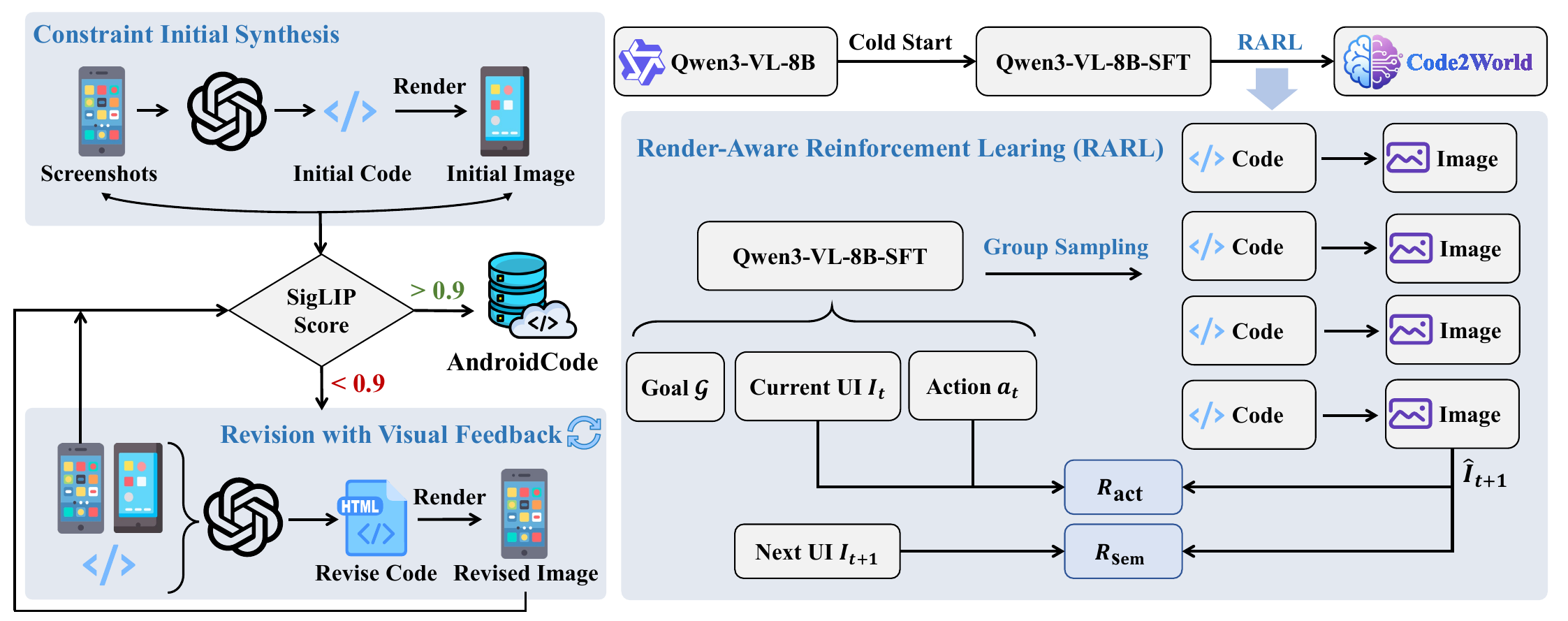}
  \caption{\textbf{Left: Illustration of Data Synthesis.} The high-fidelity \textit{AndroidCode} dataset is curated via \textit{constrainted initial synthesis} and a \textit{visual-feedback revision loop}, where synthesized HTML is iteratively refined based on rendered visual discrepancies to ensure strict alignment (SigLIP score $>$ 0.9). \textbf{Right: Two-stage Model Optimization.} The pipeline progresses from an SFT cold start to \textit{Render-Aware Reinforcement Learning (RARL)}. Utilizing Group Relative Policy Optimization (GRPO), the model optimizes dual rewards—visual semantic ($R_{\text{sem}}$) and action consistency ($R_{\text{act}}$)—derived directly from \textit{rendered outcomes} to enforce structural and logical fidelity. }
  \label{fig:pipeline}
  \vspace{-10pt}
\end{figure*}

\section{Code2World}
\label{sec:method}
We define the task as \textbf{Next UI Prediction}, which aims to approximate the deterministic state transition function of a digital environment. 
Formally, let $I_t$ denote the visual observation (screenshot) at time step $t$, $a_t$ represent the user action executed on $I_t$, and $\mathcal{G}$ denotes the task goal. The objective is to predict the subsequent visual state $\hat{I}_{t+1}$.
We propose a \textit{renderable code generation} paradigm by targeting the underlying structural representation of the interface rather than the raw pixel space. We define the state transition as a two-step conditional generation process:
\begin{equation}
    \hat{C}_{t+1} = \mathcal{M}_{\theta}(I_t, a_t, \mathcal{G}),
    \qquad
    \hat{I}_{t+1} = \mathcal{R}(\hat{C}_{t+1})
\end{equation}
where $\mathcal{M}_{\theta}$ is a multimodal generator parameterized by $\theta$, producing the predicted structured HTML code $\hat{C}_{t+1}$. 
Subsequently, the predicted visual representation $\hat{I}_{t+1}$ is obtained deterministically via a browser rendering engine $\mathcal{R}$. 

\subsection{Training Data Synthesis}
A primary challenge in training code-based world models is the scarcity of paired GUI trajectories and their corresponding clean HTML representations. Most existing benchmarks, such as Android Control~\citep{androidcontrol}, solely provide raw screenshots and action coordinates. To address this, we synthesize \textbf{AndroidCode}, a large-scale, high-fidelity dataset derived from the Android Control corpus. We implement a fully automated pipeline, visualized in the left panel of Figure \ref{fig:pipeline}, that transforms raw GUI pixels into semantic HTML code through two hierarchical stages.

\textbf{Constrained Initial Synthesis.}
We first utilize GPT-5 to translate each GUI screenshot into structured HTML code, imposing rigorous constraints to ensure deterministic rendering. Specifically, we enforce a standardized root container with fixed dimensions to align the DOM coordinate system precisely with the original screenshot. Furthermore, to mitigate the hallucination of external assets, we implement a \textit{semantic placeholder strategy}: the model is instructed to replace unreliable image URLs with descriptive text blocks (\eg~\texttt{[IMG: Red Shoe]}) and render UI icons as inline SVGs. This design guarantees that the generated HTML is a self-contained, dependency-free atomic unit. Full prompts are detailed in Appendix~\ref{appendix:data_synthesis}.

\textbf{Revision with Visual Feedback.}
Direct code synthesis from screenshots, lacking a verification loop, faces challenges in guaranteeing strict structural alignment and pixel-level precision. To ensure high-quality data curation with computational efficiency, we introduce a \textit{selective revision mechanism with visual feedback}. 
For each generated sample, we render the code back into an image and compute a visual alignment score using SigLIP against the ground truth. 
Samples exceeding a high confidence threshold are considered qualified and retained immediately. For samples falling below this threshold, we trigger a refinement loop as shown in Figure \ref{fig:pipeline}.
Specifically, we feed the ground truth screenshot, the currently rendered image, and the generated HTML code back into GPT-5. The model is instructed to visually compare the two images to identify discrepancies and rectify the code. This mechanism effectively forces the model to revise its own output based on visual evidence, ensuring that {AndroidCode} maintains high fidelity while optimizing generation efficiency.

\subsection{Model Optimization}
To empower the multimodal generator $\mathcal{M}_\theta$ with both syntactic precision and visual-level foresight, we employ a two-stage training strategy, schematically shown in the right panel of Figure \ref{fig:pipeline}. We organize the synthesized AndroidCode dataset into multimodal training triplets $(x, C^*, I^*_{t+1})$, where $x = (I_t, a_t, \mathcal{G})$ represents the conditioning input comprising the visual observation, user action, and task goal (detailed in Appendix \ref{appendix:instruction_following}). The terms $C^*$ and $I^*_{t+1}$ denote the ground-truth HTML sequence and the corresponding target GUI screenshot, respectively. This dual-modal supervision allows the model to learn from both discrete symbolic structures and rendered visual outcomes.

\textbf{Stage 1: Supervised Fine-tuning for Cold Start.}
The initial phase focuses on the symbolic mapping from $x$ to $C^*$, aiming to instill the model with the fundamental syntax of HTML and the logic of UI layout. We adapt the Qwen3-VL-8B-Instruct \cite{qwen3-vl} as backbone by minimizing the standard negative log-likelihood loss over the target code tokens $C^*$.
Through this process, the model learns to generate syntactically valid code structures. However, since this stage treats the code merely as text, the supervised policy, denoted as $\pi_{\text{sft}}$, remains blind to the final rendered visual outcome.

\textbf{Stage 2: Render-Aware Reinforcement Learning.}
To bridge the gap between textual code and visual reality, we use the final rendered outcome as the learning signal to guide optimization.
A core innovation of our framework is the design of a composite reward function $R_{total}$ that guides the model across two distinct dimensions: visual semantic fidelity and action consistency. 
Let $\hat{I}_{t+1} = \mathcal{R}(y)$ be the image rendered from a sampled HTML sequence $y$, and $I^*_{t+1}$ be the ground-truth next state.
By leveraging the reasoning capabilities of VLM, we introduce two complementary components of rewards.

\textbf{Visual Semantic Reward ($R_{sem}$).} 
Standard similarity metrics (e.g., CLIP \cite{clip}) are often brittle—overly sensitive to pixel-level shifts while missing fine-grained details~\citep{ui2code_n}. 
Furthermore, our code generation employs a \textit{rendering abstraction} strategy, where images and complex icons are represented by textual placeholders (e.g., a generic box labeled ``Settings'') since generating raw assets is infeasible. 
Although the resulting layout is semantically correct and more concise, embedding-based models severely penalize this visual style discrepancy.
To address this, we employ a VLM-as-a-Judge to evaluate the high-level semantic alignment (detailed in Appendix \ref{appendix:evaluation}).
The VLM is instructed to focus on structural layout and element correspondence, tolerating stylistic abstractions, yielding a score:
\begin{equation}
R_\text{sem} = \text{VLM}_{\text{judge}}(\hat{I}_{t+1}, I^*_{t+1})
\end{equation}
This ensures the model is rewarded for reconstructing the correct UI structure rather than superficial pixel matching.

\begin{figure*}[t]
\vspace{-8pt}
  \centering
  \includegraphics[width=\textwidth]{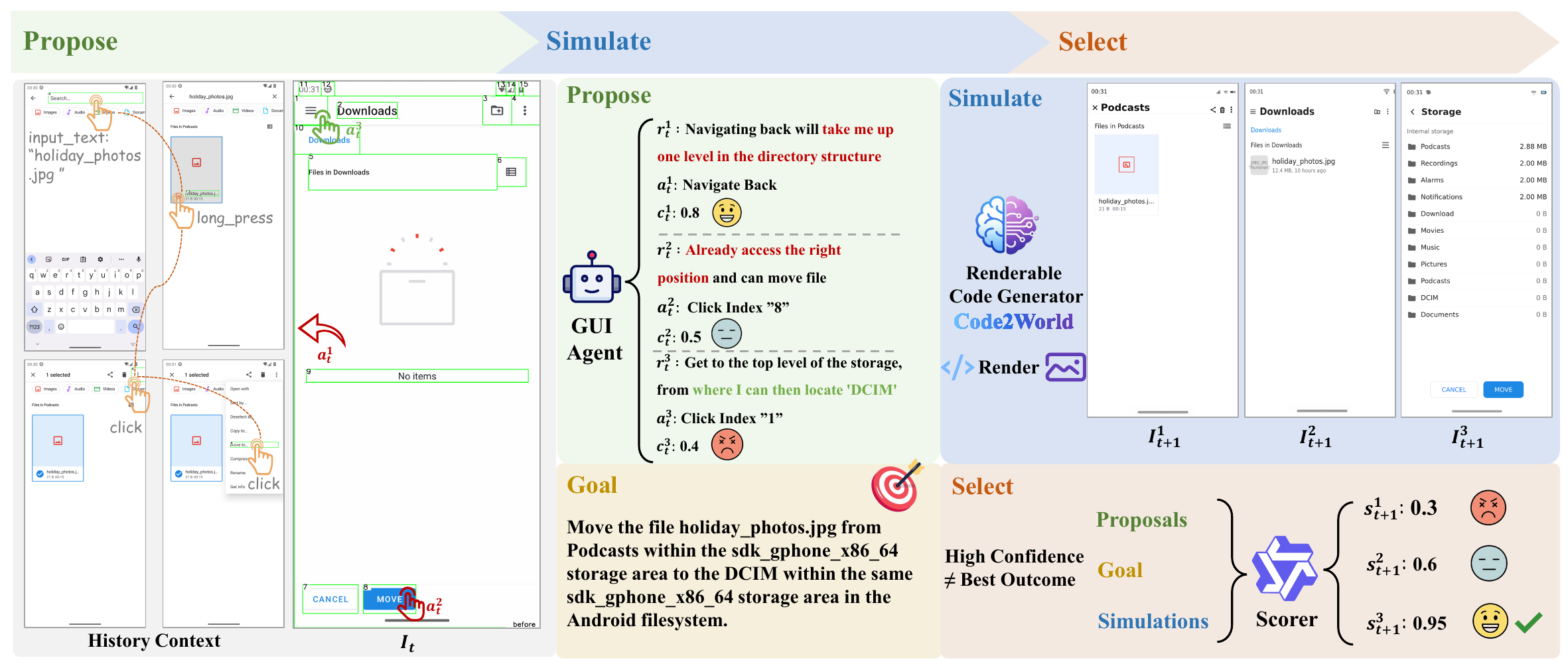}
  \caption{Illustration of the \textbf{``Propose, Simulate, Select''} pipeline for Code2World enhanced GUI agent, exemplified by an AndroidWorld task \cite{androidworld}.
  \textbf{(1) Propose}: The GUI agent generates $K$ candidate actions, with \textbf{red} and \textbf{green} highlighting hallucinated/irrational reasoning and logically sound reasoning, respectively. 
  \textbf{(2) Simulate}: Code2World predicts the execution result of each candidate via renderable code generation. 
  \textbf{(3) Select}: By evaluating the rendered future states, the system identifies the potential failure in the original policy and rectifies the decision, ultimately selecting the optimal action that aligns with the user's intent.}
  \label{fig:framework}
  \vspace{-10pt}
\end{figure*}

\textbf{Action Consistency Reward ($R_{act}$).} 
Visual similarity alone is insufficient; the state transition must logically reflect the execution of the user action $a_t$. 
We similarly utilize a VLM-as-a-Judge mechanism to verify the dynamic logic, ensuring the transition adheres to the user's intent (detailed in Appendix \ref{appendix:reward_design}).
We feed the triplet $(I_t, a_t, \hat{I}_{t+1})$ into the judge model to assess whether the state change in $\hat{I}_{t+1}$ is a valid consequence of executing $a_t$ on the previous state $I_t$. 
The VLM outputs a confidence score serving as the logic reward:
\begin{equation}
R_\text{act} = \text{VLM}_{\text{judge}}(I_t, a_t, \hat{I}_{t+1})
\end{equation}
This effectively penalizes hallucinations where the visual update contradicts the intended action logic.

The final reward is a weighted sum $R_\text{total} = \lambda_1 R_\text{sem} + \lambda_2 R_\text{act}$, where $\lambda_1$ and $\lambda_2$ balance each component’s contribution.
To optimize this objective, we adopt \textit{Group Relative Policy Optimization (GRPO)} \cite{grpo}, which estimates the baseline using the group average of rewards from $G$ sampled outputs $\{y_1, \dots, y_G\}$ from the old policy $\pi_{\theta_{\text{old}}}$. 
For each input $x$, we compute the advantages $A_i$ by normalizing rewards within the group:
\begin{equation}
A_i = \frac{R_\text{total}(y_i) - \text{mean}(\{R_\text{total}(y_j)\}_{j=1}^G)}{\text{std}(\{R_\text{total}(y_j)\}_{j=1}^G) + \epsilon}
\end{equation}

The model parameters $\theta$ are then updated by maximizing the surrogate objective with a KL-divergence penalty to maintain proximity to the reference SFT policy:

\begin{equation}
\small
\begin{aligned}
    \mathcal{L}_{\text{GRPO}}(\theta) = \mathbb{E}_{x} \biggl[ \frac{1}{G} \sum_{i=1}^{G} \min \bigl( \rho_i A_i, \text{clip}(\rho_i, 1-\epsilon, \\ 1+\epsilon) A_i \bigr) - \beta D_{\text{KL}}(\pi_\theta \parallel \pi_{\text{sft}}) \biggr],
\end{aligned}
\label{eq:grpo}
\end{equation}

where $\rho_i=\frac{\pi_\theta(\mathbf{y}_i|\mathbf{x})}{\pi_{\theta_{\text{old}}}(\mathbf{y}_i|\mathbf{x})} $ is the importance sampling ratio. Through this process, Code2World iteratively refines its generation policy to maximize both visual fidelity and logical coherence.

\section{Evaluation and Application of Code2World}
What defines a high-quality GUI world model? 
We posit that a reliable virtual sandbox in GUI domain must satisfy two critical requirements:
(1) \textbf{Precise Next UI Prediction}, where the model must simultaneously achieve high visual fidelity and strictly adhere to interaction logic during state transitions, and 
(2) \textbf{Effective GUI Agent Enhancement}, where the world model's foresight effectively enhances GUI agent planning and decision-making. 
Accordingly, we structure our evaluation to benchmark simulation quality in Section \ref{sec:eval_next_ui_pred}, followed by an assessment of Code2World's practical impact on assisting GUI agents in Section \ref{sec:enhance_gui_agent}.

\subsection{Evaluation for Next UI Prediction}
\label{sec:eval_next_ui_pred}
To comprehensively assess the capability of next UI prediction, we propose a holistic evaluation protocol tailored for GUI environments. Unlike general image generation tasks, GUI dynamics requires strict adherence to interaction rules and precise structural rendering. Therefore, we design metrics across two complementary dimensions by employing a unified VLM-as-a-Judge framework.

\textbf{Functional Logic.} 
This dimension evaluates the correctness of state transitions, verifying whether the simulation adheres to the underlying interaction rules.
\begin{itemize}[leftmargin=*, topsep=0pt, partopsep=0pt, itemsep=2pt, parsep=0pt]

    \item \textbf{Action Adherence $S_{ad}$}: 
    Determines if the generated state logically ensues from the executed action and the initial state, preventing hallucinations where the visual update contradicts the intended interaction logic.

    \item \textbf{Action Identifiability $S_{id}$}: 
    Tests if the executed action can be correctly inferred solely from the visual difference between the origin and the generated state, measuring the causal clarity of the simulation.
\end{itemize}

\textbf{Visual Quality.} 
This dimension assesses the fidelity of the rendered interface. Beyond standard semantic metrics like SigLIP \cite{siglip} and DINO \cite{dinov2} which capture high-level similarity, we introduce two specialized metrics to scrutinize fine-grained details:

\begin{itemize}[leftmargin=*, topsep=0pt, partopsep=0pt, itemsep=2pt, parsep=0pt]

    \item \textbf{Element Alignment $S_{ele}$}:
    Checks if key UI elements align precisely with the ground truth layout.

    \item \textbf{Layout Integrity $S_{lay}$}: 
    Evaluates the preservation of layout geometry and structural containment.
\end{itemize}
The formal formulations and detailed implementations of these metrics are provided in Appendix \ref{appendix:eval_metrics}.


\subsection{Application for GUI Agent}
\label{sec:enhance_gui_agent}
Code2World serves as a plug-and-play module to enhance any existing GUI agent. 
As illustrated in Figure \ref{fig:framework}, we seamlessly integrate it as a look-ahead simulator during inference via \textbf{``Propose, Simulate, Select''} pattern.

Given the current observation $I_t$, the task goal $\mathcal{G}$ and the interaction history $\mathcal{H}_t$, the GUI agent generates a set of $K$ candidate proposals $\mathcal{P}_t$, where each proposal comprises a reasoning rationale $r_t^{(k)}$, an executable action $a_t^{(k)}$, and an initial confidence score $c_t^{(k)}$:
\begin{equation}
\mathcal{P}_t = \text{Agent}(I_t, \mathcal{G}, \mathcal{H}_t) = \left\{ \left(r_t^{(k)}, a_t^{(k)}, c_t^{(k)}\right) \right\}_{k=1}^{K}.
\end{equation}

For each proposal, Code2World predicts the next GUI state by generating structured HTML Code and rendering it into an image:
\begin{equation}
C_{t+1}^{(k)} = \mathcal{M}_\theta(I_t, a_t^{(k)}, \mathcal{G}), \quad \hat{I}_{t+1}^{(k)} = \mathcal{R}\left(C_{t+1}^{(k)}\right).
\end{equation}

Finally, the scorer selects the most promising action by evaluating the task goal and simulated outcomes:
\begin{equation}
a_t^\star = \arg\max_{a_t^{(i)} \in \mathcal{A}_t}
\; \mathcal{S}\big(I_t, \mathcal{G}, a_t^{(i)}, \hat{I}_{t+1}^{(i)}\big),
\end{equation}
where $\mathcal{S}(\cdot)$ can be implemented as a VLM-based verifier that judges which predicted next state best advances progress toward the task goal $\mathcal{G}$, effectively filtering out hallucinations or illogical plans.
\section{Experiments}
We structure our experimental analysis to investigate three core research questions:
\textbf{RQ1:} How effectively can Code2World predict next GUI observation in both in-domain and out-of-distribution settings?
\textbf{RQ2:} Can Code2World enhance the GUI agents navigation in both offline and online setting?
\textbf{RQ3:} How does each component of Code2World contribute to the overall performance?
Notably, for each setting, we carefully tailor the baselines and metrics to the corresponding research question.

\begin{table*}[h]
\vspace{-5pt}
    \centering
    \caption{\textbf{Quantitative Comparison} of various image and  code generation models on Android Control \textbf{(ID)} to assess basic capabilities on the same device, and GUI Odyssey \textbf{(OOD)} to test generalization robustness across unseen devices and cross-app scenarios. The best scores are in \textbf{bold} while the second best are in \underline{underlined}.}
    \resizebox{\textwidth}{!}{%
        \begin{tabular}{l c c c c c c c c c c c c}
            \toprule
            \multirow{3}{*}{\textbf{Model}} & 
            \multicolumn{6}{c}{\textbf{Android Control (ID)}} & 
            \multicolumn{6}{c}{\textbf{GUI Odyssey (OOD)}} \\
            \cmidrule(lr){2-7} \cmidrule(lr){8-13}
            
            & \multicolumn{2}{c}{\textbf{Functional Logic}} & \multicolumn{4}{c}{\textbf{Visual Quality}} 
            & \multicolumn{2}{c}{\textbf{Functional Logic}} & \multicolumn{4}{c}{\textbf{Visual Quality}} \\
            \cmidrule(lr){2-3} \cmidrule(lr){4-7} \cmidrule(lr){8-9} \cmidrule(lr){10-13}
            
            & $S_{ad}$ & $S_{id}$ & $S_{ele}$ & $S_{lay}$ & \textit{SigLIP} & \textit{DINO} 
            & $S_{ad}$ & $S_{id}$ & $S_{ele}$ & $S_{lay}$ & \textit{SigLIP} & \textit{DINO} \\

            \midrule
            \rowcolor{groupbg}
            \multicolumn{13}{c}{\textbf{\textit{Image Generation Models}}} \\
            \midrule
            
            Gemini-3-Pro-Image   & 92.63 & 83.67 & 68.47 & 63.67 & \textbf{84.89} & \textbf{64.11} & 88.50 & 75.20 & 45.60 & 42.30 & \textbf{78.74} & \textbf{48.10} \\
            GPT-Image-1       & 89.59 & 71.40 & 58.78 & 56.22 & 77.58 & \underline{61.00} & 91.63 & 71.40 & 42.04 & 39.80 & \underline{76.46} & \underline{46.64} \\
            Doubao-Seedream-4.5 & 85.36 & 86.15 & 59.08 & 55.82 & 81.76 & 52.19 & 89.30 & 66.04 & 42.71 & 40.75 & 72.80 & 37.83 \\
            Qwen-Image-Edit-Max   & 57.55 & 54.12 & 54.05 & 46.33 & 70.39 & 53.61 & 69.05 & 65.11 & 38.24 & 34.28 & 66.46 & 47.77 \\
            Janus-Pro-7B   & 58.10 & 53.68 & 55.22 & 45.90 & 79.85 & 54.30 & 68.40 & 63.30 & 39.10 & 33.95 & 67.05 & 48.20 \\

            \midrule
            \rowcolor{groupbg}
            \multicolumn{13}{c}{\textbf{\textit{Code Generation Models}}} \\
            \midrule
            GPT-5             & \underline{94.02} & \textbf{90.15} & \underline{74.13} & \underline{69.78} & 78.07 & 41.74 & \textbf{93.45} & \underline{78.22} & \textbf{64.41} & \textbf{60.33} & 70.79 & 46.35 \\          
            Gemini-3-Flash    & 92.65 & 84.08 & \textbf{74.52} & 69.74 & 81.17 & 45.72 & 92.61 & \textbf{82.04} & \underline{59.43} & \underline{55.12} & 71.19 & 43.27 \\
            Claude-Sonnet-4.5 & 89.60 & 86.12 & 65.80 & 62.25 & 75.97 & 36.72 & 82.81 & 79.17 & 43.75 & 41.56 & 64.72 & 34.94 \\
            JanusCoderV-7B    & 57.12 & 56.05 & 30.18 & 31.09 & 59.89 & 21.73 & 86.64 & 68.11 & 46.42 & 43.85 & 74.82 & 41.65 \\
            Qwen3-VL-8B       & 59.20 & 65.80 & 43.10 & 42.70 & 63.88 & 28.86 & 34.60 & 40.00 & 14.20 & 14.00 & 58.95 & 32.49 \\
            Qwen2.5-VL-72B    & 73.34 & 70.12 & 47.15 & 48.26 & 68.45 & 30.12 & 56.12 & 54.08 & 21.45 & 20.73 & 55.25 & 30.20 \\
            InternVL3-78B     & 72.41 & 67.35 & 45.06 & 46.73 & 62.64 & 25.42 & 60.25 & 58.10 & 25.30 & 24.88 & 60.45 & 31.50 \\
            GLM-4.6V-106B   & 91.62 & 74.23 & 61.95 & 58.74 & 67.26 & 27.42 & 55.40 & 52.80 & 30.15 & 28.90 & 62.10 & 33.25 \\

            \midrule
            \textbf{Code2World-8B (Ours)} & \textbf{94.28} & \underline{88.64} & 71.35 & \textbf{70.32} & \underline{79.44} & 49.18 & \underline{92.73} & \underline{78.22} & 55.46 & 49.78 & 74.40 & 46.43 \\
            

            \bottomrule
        \end{tabular}%
    }
    \label{tab:main}
    \vspace{-15pt}
\end{table*}

\begin{figure*}[t]
\vspace{-8pt}
  \centering
\includegraphics[width=\textwidth]{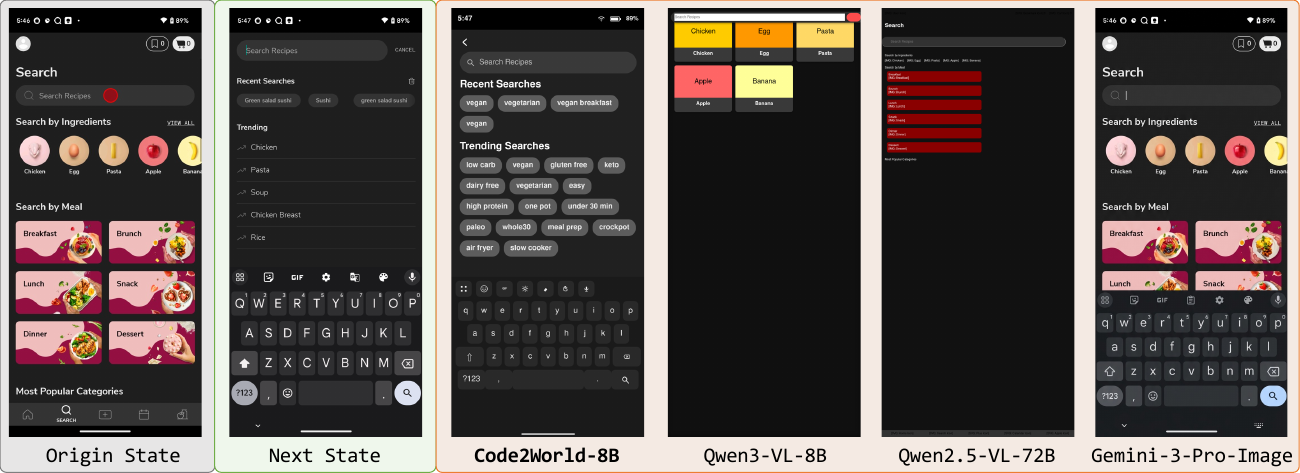}
  \caption{\textbf{Qualitative comparison} of next GUI state generation over Code2World and three baselines. The red circle in origin state indicates the user's click position targeting the search bar.}
  \vspace{-5pt}
  \label{fig:show_case}
\end{figure*}

\subsection{World Model Ability (RQ1)}

\textbf{Benchmarks and Metrics.} 
We rigorously evaluate the generalization capability of Code2World on unseen applications. We employ two benchmarks: (1) \textbf{Android Control} serves as the \textbf{In-Domain (ID)} setting, containing applications not seen during training, evaluating generalization within the same mobile device used in training. (2) \textbf{GUI Odyssey} represents the \textbf{Out-of-Distribution (OOD)} setting, serving as a more challenging \textit{Cross-App} setting with diverse UI styles and domains to test robustness against out-of-distribution device shifts. 
To strictly quantify performance, we report results using the four VLM-based metrics ($S_{ad}$, $S_{id}$, $S_{ele}$, $S_{lay}$) defined in our evaluation protocol (Section \ref{sec:eval_next_ui_pred}), along with standard image similarity metrics (SigLIP, DINO).

\textbf{Baselines.}
We compare Code2World against a comprehensive suite of state-of-the-art models, categorized by their generation modality. 
For \textbf{Image Generation Models}, we compare against pixel-space synthesis approaches, specifically Gemini-3-Pro-Image \cite{gemini3pro}, GPT-Image-1, Doubao-Seedream-4.5, Qwen-Image-Edit-Max, and Janus-Pro-7B \cite{janus-pro}.
For \textbf{Code Generation Models}, we evaluate VLMs in a zero-shot setting where they predict the next GUI state by directly generating HTML. This includes proprietary models such as Claude-4.5-Sonnet, Gemini-3-Flash \cite{gemini2.5}, and GPT-5 \cite{gpt-4o}, alongside leading open-source models including JanusCoderV-7B \cite{januscoder}, Qwen3-VL-8B \cite{qwen3-vl}, Qwen2.5-VL-72B \cite{qwen2.5-vl}, InternVL3-78B \cite{internvl3}, and GLM-4.6V-106B \cite{glm4.6}. 




\textbf{Quantitative Comparison.}
As shown in Table \ref{tab:main}, Code2World is \textbf{lightweight yet powerful}. Despite its compact 8B size, Code2World outperforms open-source baselines scaling \textbf{over 10 $\times$} in parameters across both dynamic logic and visual quality dimensions, rivaling proprietary giants like {GPT-5} and {Gemini-3-Pro-Image}.
Our analysis identifies distinct limitations inherent to existing \textit{open-source approaches}. 
Large generalist VLMs (e.g., InternVL3-78B, GLM-4.6V-106B), while capable of inferring semantic state transitions, often lack the specialized UI-to-Code alignment required to faithfully reconstruct the future layout, evidenced by their suboptimal visual quality scores. 
Conversely, image editing models (e.g., Qwen-Image-Edit, Janus-Pro-7B) struggle to predict precise interface dynamics, prioritizing local texture consistency over the global interaction logic necessary for navigation. 
Code2World effectively bridges these gaps, delivering high-fidelity simulation in both structure and logic without such shortcomings. This validates the potential of \textit{renderable code generation} paradigm to unlock  GUI world modeling capabilities in lightweight VLM architectures.

\textbf{Robust Generalization.} 
Performance on the GUI Odyssey benchmark (OOD) further validates Code2World’s capability to internalize interaction dynamics rather than merely memorizing specific layouts.
As expected, the shift to diverse, unseen cross-app environments causes a natural decline in visual similarity metrics across all models compared to the in-domain Android Control.
Crucially, however, Code2World exhibits exceptional robustness in dynamic logic, \textbf{maintaining  $S_{ad}$ of 92.73} and \textbf{$S_{id}$ of 78.22}—minimal fluctuation relative to the marked performance decay seen in open-source counterparts.
This consistency suggests that Code2World has internalized the fundamental interaction dynamics of GUIs, which confirm its viability as a general-purpose world model capable of operating in novel digital environments.

\begin{table}[h]
    \centering
    \caption{Performance comparison on AndroidControl-High.}
    \small

    
    
    \begin{tabularx}{\linewidth}{
        >{\raggedright\arraybackslash}p{0.40\linewidth}
        >{\centering\arraybackslash}p{0.13\linewidth}
        >{\centering\arraybackslash}p{0.13\linewidth}
        >{\centering\arraybackslash}p{0.13\linewidth}
    }

        \toprule
        \multirow{2}{*}{\textbf{Model}} & \multicolumn{3}{c}{\textbf{AndroidControl-High}} \\
        \cmidrule(lr){2-4}
         & Type & Grounding & SR \\
        \midrule
        
        GPT-4o             & 62.14 & 31.82 & 21.2 \\
        Gemini-2.5-Flash  & 67.43 & 33.29 & 27.9 \\
        GUI-R1-7B         & 78.45 & 75.64 & 67.15 \\
        InfiGUI-R1-3B     & 83.16  & 74.51  & 70.98  \\
        UI-TARS-1.5-7B     & 73.36 & 77.02 & 61.57 \\
        
        \midrule 
        
        Mobile-Agent-v3-7B     & 82.05 & 75.16 & 67.20 \\
        \hspace{0.5em}\textbf{+Code2World} & 84.13 & 78.78 & 68.41 \\ 
        
        
        \rowcolor{impbg}
        \makebox[0pt][l]{\color{imptext}$\Delta$ \textbf{\textit{Improvement}}} &
        \makebox[0pt][c]{\color{imptext}\textbf{\textit{+2.09}}} &
        \makebox[0pt][c]{\color{imptext}\textbf{\textit{+3.62}}} &
        \makebox[0pt][c]{\color{imptext}\textbf{\textit{+1.21}}} \\

        \midrule 
        
        Qwen2.5-VL-7B     & 76.66 & 69.64 & 65.16 \\
        \hspace{0.5em}\textbf{+Code2World} & 78.74 & 74.87 & 66.47 \\ 
        
        \rowcolor{impbg} 
        \makebox[0pt][l]{\color{imptext}$\Delta$ \textbf{\textit{Improvement}}} & 
        \makebox[0pt][c]{\color{imptext}\textbf{\textit{+2.08}}} & 
        \makebox[0pt][c]{\color{imptext}\textbf{\textit{+5.23}}} & 
        \makebox[0pt][c]{\color{imptext}\textbf{\textit{+1.31}}} \\
                
        \bottomrule
    \end{tabularx}
    \vspace{-5pt}
    \label{tab:ac}
\end{table}

\begin{figure}[t]
\vspace{-8pt}
  \centering
  \includegraphics[width=\columnwidth]{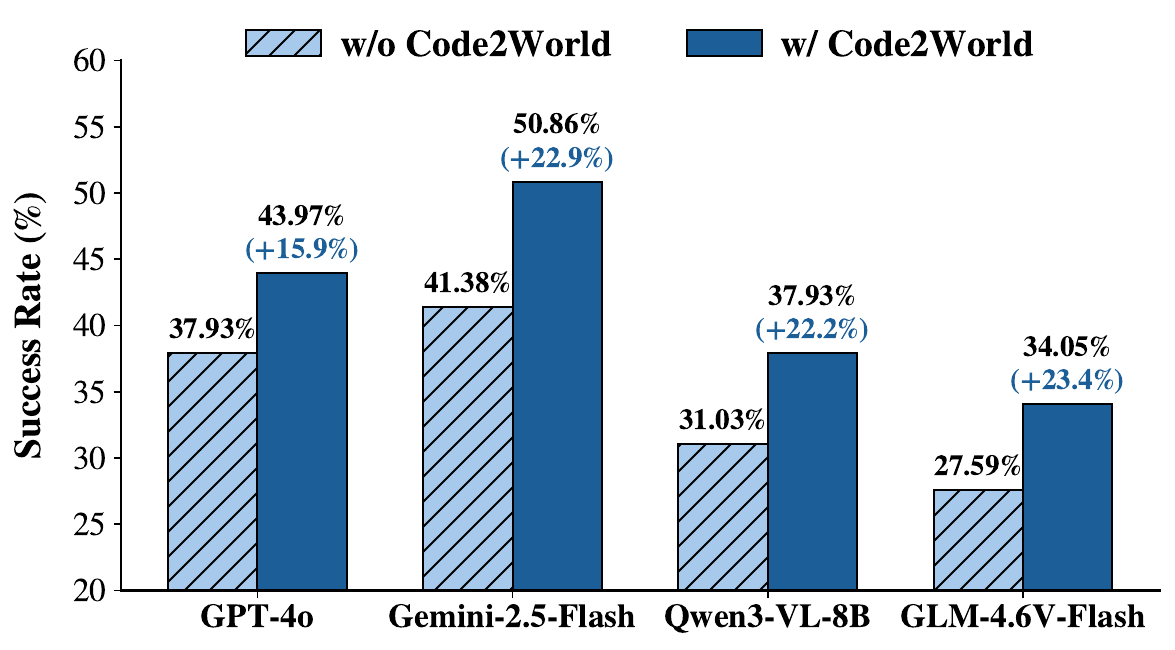}
  \caption{Performance comparison on the AndroidWorld.}
  \label{fig:aw}
  \vspace{-15pt}
\end{figure}

\textbf{Qualitative Comparison.}
Figure \ref{fig:show_case} provides a visual confirmation that directly corroborates the failure modes identified in our quantitative analysis. Reflecting the capacity deficit of smaller models, Qwen3-VL-8B fails to comprehend the transition logic entirely, merely reproducing the initial state with artifacts rather than predicting the future. Exemplifying the reasoning-execution gap, Qwen2.5-VL-72B correctly infers the semantic intent to switch to a ``search'' interface but fails to translate this into a coherent visual form, resulting in a chaotic layout that lacks clarity. Similarly, verifying the structural rigidity of pixel-based generation, Gemini-3-Pro-Image excels at texture synthesis but remains constrained to the original layout geometry, unable to hallucinate the entirely new page structure required by the navigation. 
In stark contrast, {Code2World} bridges these gaps by accurately predicting both the logical jump and fine-grained visual details, achieving {near-perfect alignment} with the ground truth.
A compelling instance is that Code2World even \textbf{simulates temporal passage}, evidenced by the time change ($5:46\rightarrow 5:47$) during the state transition.
This intriguing detail exemplifies the model's robust world modeling capabilities, demonstrating its aptitude for capturing latent environmental changes and predicting them with both logical correctness and high visual fidelity.
More Qualitative Analysis can be seen in Appendix \ref{appendix:visualization_world_modeling}.

\subsection{GUI Agent Enhancement (RQ2)}

\textbf{Offline Navigation.}
We evaluate the effectiveness of Code2World in enhancing agents’ capabilities on AndroidControl-High~\cite{androidcontrol}. This benchmark is highly challenging, as it provides only the user task, requiring the agent to autonomously plan step-wise action, thereby evaluating its single-step decision-making capability. We apply Code2World to enhance both a specialized GUI agent (Mobile-Agent-v3~\cite{mobileagentv3}) and a general MLLM (Qwen2.5-VL-7B), and further compare against GPT-4o~\cite{gpt-4o}, Gemini-2.5-Flash~\cite{gemini2.5}, GUI-R1-7B~\cite{gui-r1}, InfiGUI-R1-3B~\cite{infigui-r1}, and UI-TARS-1.5-7B~\cite{ui-tars}.  We report \textit{Action Type} accuracy, \textit{Grounding Accuracy}, and the overall \textit{Success Rate (SR)}.

As shown in Table \ref{tab:ac}, incorporating Code2World yields a substantial improvement for the Qwen2.5-VL-7B, achieving an improvement of 5.23 in Grounding accuracy, making it comparable to domain-specific GUI agents and even surpassing InfiGUI-R1. Moreover, although Mobile-Agent-v3-7B is trained on task-specific data and already attains strong performance, it still benefits from Code2World, achieving the best accuracy on both Type and Grounding, which validates the plug-and-play versatility of our approach. 
In summary, by functioning as a \textit{model-agnostic simulator} to enable future GUI prediction, Code2World empowers the agent to better understand the consequences of each action, thereby selecting actions that more closely align with user intent and significantly improving single-step decision-making performance.


\textbf{Online Application.}
Although Code2World demonstrates effective improvements in the offline setting, such evaluation is inherently constrained by pre-recorded human trajectories. As a result, it cannot assess critical capabilities required in real-world deployment, such as exploring alternative valid trajectories or recovering from errors. To further validate the effectiveness of Code2World in real-world navigation scenarios, we conduct online evaluation in AndroidWorld~\cite{androidworld}, an environment designed for developing and benchmarking autonomous agents on a live Android emulator, comprising 116 tasks across 20 mobile applications. Unlike offline evaluation, which focuses on single-step action prediction, AndroidWorld requires agents to generate continuous action sequences and measures performance by the task success rate (SR). We adopt the default M3A agent framework provided by AndroidWorld and evaluate multiple VLMs, including two closed-source models (GPT-4o, Gemini-2.5-Flash) and two open-source models (Qwen3-VL-8B, GLM-4.6V-Flash \cite{glm4.6}). 

As shown in Figure~\ref{fig:aw}, Code2World consistently improves task success rates across all evaluated models, demonstrating its plug-and-play effectiveness. This improvement stems from Code2World’s renderable code generation and render-aware reinforcement learning strategy, which enable accurate and realistic GUI prediction. By providing agents with foresight into future GUI states, Code2World allows effective exploration of multiple candidate actions while reliably selecting the most advantageous one, thereby substantially enhancing long-horizon reasoning ability in online interaction.
More Visualization cases of Code2World enhance GUI Agent decision-making can be found in the Appendix \ref{appendix:visualization_enhance_gui_agent}.

\begin{table}[t]
    \centering
    \caption{Performance comparison world model ability of Qwen3-VL-8B (Base) and Code2World variants on Android Control.}
    \resizebox{\linewidth}{!}{%
    \begin{tabular}{l c c c c c c}
        \toprule
        \multirow{3}{*}{\textbf{Model}} 
        & \multicolumn{6}{c}{\textbf{Android Control}} \\
        \cmidrule(lr){2-7}

        & \multicolumn{2}{c}{\textbf{Functional Logic}} 
        & \multicolumn{4}{c}{\textbf{Visual Quality}} \\
        \cmidrule(lr){2-3} \cmidrule(lr){4-7}

        & $S_{ad}$ & $S_{id}$ & $S_{ele}$ & $S_{lay}$ & \textit{SigLIP} & \textit{DINO} \\
        \midrule

        Qwen3-VL-8B           & 59.20 & 65.80 & 43.10 & 42.70 & 63.88 & 28.86 \\
        Qwen3-VL-8B$+$SFT     & 78.45 & 79.12 & 58.65 & 56.30 & 71.50 & 39.40 \\
        Qwen3-VL-8B$+$SFT$+$$R_{sem}$ & 78.90 & 78.85 & \underline{65.40} & \underline{61.20} & \underline{76.10} & \underline{44.50} \\
        Qwen3-VL-8B$+$SFT$+$$R_{act}$ & \underline{88.20} & \underline{84.53} & 60.10 & 57.80 & 74.80 & 40.60 \\
        \textbf{Code2World} & \textbf{94.28} & \textbf{88.64} & \textbf{71.35} & \textbf{70.32} & \textbf{79.44} & \textbf{49.18} \\

        \bottomrule
    \end{tabular}%
    }
    \label{tab:android_control_ablation}
    \vspace{-10pt}
\end{table}

\subsection{Ablation Study (RQ3)}

\textbf{Impact of Components on Next UI Prediction.} 
We investigate the contribution of each training stage in Table~\ref{tab:android_control_ablation}. 
Applying SFT instills fundamental HTML syntax and layout rules, establishing a strong foundation that yields substantial improvements across both functional logic and visual quality metrics. 
However, refining the model solely with the visual reward ($R_{sem}$) reveals a limitation: while rendering quality improves, functional logic remains stagnant or slightly regresses. 
This suggests a degree of reward hacking, where the model prioritizes superficial pixel alignment with the ground truth rather than mastering the underlying state transition logic essential for a world model. 
Conversely, the action reward ($R_{act}$) primarily boosts dynamic logic by enforcing correct state transitions. 
While this ensures the generated view corresponds to the correct target state, resulting in moderate visual gains, it lacks the fine-grained feedback needed for high-fidelity rendering. 
Ultimately, Code2World integrates both rewards to achieve optimal performance, verifying that combining semantic alignment with interaction logic is critical for a robust world model.

\begin{figure}[t]
  \centering
  \includegraphics[width=\columnwidth]{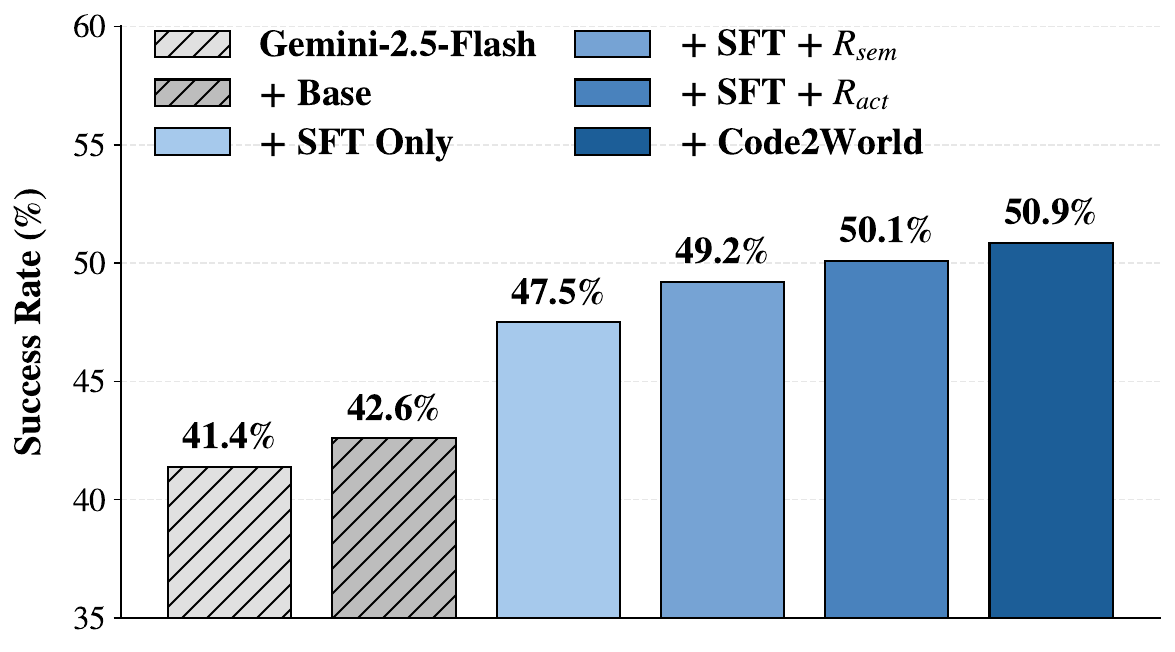}
  \caption{Ablation analysis of training pipeline design for task success rate (SR) on AndroidWorld. ``Base'' denotes Qwen3-VL-8B. ``+SFT only'' and reward-based variants (``+SFT+ $R_{\mathrm{sem}}$'', ``+SFT+ $R_{\mathrm{act}}$'') represent additional training applied on the Base model.} 
  \label{fig:ablation_aw}
  \vspace{-15pt}
\end{figure}

\textbf{Impact of Components on GUI Agent Enhancement.} 
We further analyze how varying qualities of the world model impact the downstream decision-making of the Gemini-2.5-Flash agent on the AndroidWorld benchmark. As shown in Figure~\ref{fig:ablation_aw}, the standalone Gemini-2.5-Flash agent achieves a baseline Success Rate (SR) of 41.4\%. Merely equipping it with a naive simulator (+ Qwen3-VL-8B) yields a negligible gain (+1.2\%), indicating that low-fidelity predictions fail to provide reliable foresight. In contrast, integrating our SFT model triggers a sharp performance jump to 47.5\%, demonstrating that structured, parsable HTML predictions effectively ground the agent's planning. The subsequent RL stages progressively refine this capability: $R_{sem}$ enhances the visual clarity of the simulated future, pushing the SR to 49.2\%, while $R_{act}$ ensures the simulator correctly reflects interaction dynamics, further raising the SR to 50.1\%. Finally, the full Code2World model, combining all optimizations, achieves a peak SR of \textbf{50.9\%} (\textbf{+9.5\%} improvement over the baseline). This strong correlation between world model fidelity and agent success rate underscores that accurate, renderable visual foresight is the key to unlocking robust long-horizon reasoning.

\section{Conclusion}

In this work, we introduced Code2World, a pioneering code-native GUI world model that fundamentally shifts next UI prediction from raw pixel estimation to \textit{renderable HTML code generation}, uniquely combining high-fidelity visualization with fine-grained structural controllability.
We constructed AndroidCode (over 80K screen--action pairs) using a visual-feedback revision loop, and proposed {Render-Aware Reinforcement Learning} to align code prediction with rendered visual fidelity and action consistency. 
Functioning as a learnable virtual sandbox, Code2World empowers autonomous GUI agents to navigate complex, dynamic interfaces with human-like foresight. 
Empirically, Code2World-8B achieves state-of-the-art performance in next UI prediction and, acting as a plug-and-play simulator, significantly enhances downstream agents, boosting Gemini-2.5-Flash by +9.5\% on AndroidWorld navigation.
\section*{Impact Statement}
This paper presents a significant advancement in the field of autonomous GUI agents. The primary societal benefit of our work lies in enhancing digital inclusivity. By empowering agents to handle complex interface interactions, this technology serves as an assistive tool for users with disabilities while simultaneously relieving humans from repetitive digital labor. Uniquely, as a World Model, Code2World contributes to AI safety by providing a sandbox environment. This allows agents to simulate and evaluate potentially irreversible actions, such as financial transactions or data deletion, without executing them in the real world, thereby mitigating the risks associated with on-policy trial-and-error learning. However, we acknowledge potential risks. If the world model hallucinates incorrect safety cues, it could mislead an agent into executing harmful actions. Additionally, as with any advanced automation technology, there is a potential for misuse in automated cyber-attacks or navigation spamming. We encourage the research community to focus on robust verification mechanisms to ensure that the predictive capabilities of such models are deployed responsibly and ethically.

\nocite{langley00}

\bibliography{example_paper}

@inproceedings{showui,
  title={Showui: One vision-language-action model for gui visual agent},
  author={Lin, Kevin Qinghong and Li, Linjie and Gao, Difei and Yang, Zhengyuan and Wu, Shiwei and Bai, Zechen and Lei, Stan Weixian and Wang, Lijuan and Shou, Mike Zheng},
  booktitle={Proceedings of CVPR},
  year={2025}
}

@article{ui-tars,
  title={Ui-tars: Pioneering automated gui interaction with native agents},
  author={Qin, Yujia and Ye, Yining and Fang, Junjie and Wang, Haoming and Liang, Shihao and Tian, Shizuo and Zhang, Junda and Li, Jiahao and Li, Yunxin and Huang, Shijue and others},
  journal={arXiv preprint arXiv:2501.12326},
  year={2025}
}

@inproceedings{cogagent,
  title={Cogagent: A visual language model for gui agents},
  author={Hong, Wenyi and Wang, Weihan and Lv, Qingsong and Xu, Jiazheng and Yu, Wenmeng and Ji, Junhui and Wang, Yan and Wang, Zihan and Dong, Yuxiao and Ding, Ming and others},
  booktitle={Proceedings of CVPR},
  year={2024}
}

@inproceedings{mga,
  title={MGA: Memory-Driven GUI Agent for Observation-Centric Interaction},
  author={Cheng, Weihua and Ni, Ersheng and Wang, Wenlong and Sun, Yifei and Liu, Junming and Shen, Wangyu and Chen, Yirong and Shi, Botian and Wang, Ding},
  booktitle={Proceedings of WSDM},
  year={2025}
}

@inproceedings{gui-rise,
  title={GUI-Rise: Structured Reasoning and History Summarization for GUI Navigation},
  author={Liu, Tao and Wang, Chongyu and Li, Rongjie and Yu, Yingchen and He, Xuming and Song, Bai},
  booktitle={Proceedings of NIPS},
  year={2025}
}

@inproceedings{gui-reflection,
  title={GUI-Reflection: Empowering Multimodal GUI Models with Self-Reflection Behavior},
  author={Wu, Penghao and Ma, Shengnan and Wang, Bo and Yu, Jiaheng and Lu, Lewei and Liu, Ziwei},
  booktitle={Proceedings of NIPS},
  year={2025}
}

@inproceedings{ui-genie,
  title={UI-Genie: A Self-Improving Approach for Iteratively Boosting MLLM-based Mobile GUI Agents},
  author={Xiao, Han and Wang, Guozhi and Chai, Yuxiang and Lu, Zimu and Lin, Weifeng and He, Hao and Fan, Lue and Bian, Liuyang and Hu, Rui and Liu, Liang and others},
  booktitle={Proceedings of NIPS},
  year={2025}
}

@article{mobileagentv3,
  title={Mobile-Agent-v3: Foundamental Agents for GUI Automation},
  author={Ye, Jiabo and Zhang, Xi and Xu, Haiyang and Liu, Haowei and Wang, Junyang and Zhu, Zhaoqing and Zheng, Ziwei and Gao, Feiyu and Cao, Junjie and Lu, Zhengxi and others},
  journal={arXiv preprint arXiv:2508.15144},
  year={2025}
}

@article{gui-r1,
  title={Gui-r1: A generalist r1-style vision-language action model for gui agents},
  author={Luo, Run and Wang, Lu and He, Wanwei and Chen, Longze and Li, Jiaming and Xia, Xiaobo},
  journal={arXiv preprint arXiv:2504.10458},
  year={2025}
}

@article{ui-r1,
  title={UI-R1: Enhancing Efficient Action Prediction of GUI Agents by Reinforcement Learning},
  author={Lu, Zhengxi and Chai, Yuxiang and Guo, Yaxuan and Yin, Xi and Liu, Liang and Wang, Hao and Xiao, Han and Ren, Shuai and Xiong, Guanjing and Li, Hongsheng},
  journal={arXiv preprint arXiv:2503.21620},
  year={2025}
}

@article{infigui-r1,
  title={Infigui-r1: Advancing multimodal gui agents from reactive actors to deliberative reasoners},
  author={Liu, Yuhang and Li, Pengxiang and Xie, Congkai and Hu, Xavier and Han, Xiaotian and Zhang, Shengyu and Yang, Hongxia and Wu, Fei},
  journal={arXiv preprint arXiv:2504.14239},
  year={2025}
}

@article{mobilegui-rl,
  title={MobileGUI-RL: Advancing Mobile GUI Agent through Reinforcement Learning in Online Environment},
  author={Shi, Yucheng and Yu, Wenhao and Li, Zaitang and Wang, Yonglin and Zhang, Hongming and Liu, Ninghao and Mi, Haitao and Yu, Dong},
  journal={arXiv preprint arXiv:2507.05720},
  year={2025}
}

@inproceedings{arpo,
  title={ARPO: End-to-End Policy Optimization for GUI Agents with Experience Replay},
  author={Lu, Fanbin and Zhong, Zhisheng and Liu, Shu and Fu, Chi-Wing and Jia, Jiaya},
  booktitle={Proceedings of NIPS},
  year={2025}
}

@article{ui2code_n,
  title={{UI2Code}$^{\text{N}}$: A Visual Language Model for Test-Time Scalable Interactive UI-to-Code Generation},
  author={Yang, Zhen and Hong, Wenyi and Xu, Mingde and Fan, Xinyue and Wang, Weihan and Cheng, Jiele and Gu, Xiaotao and Tang, Jie},
  journal={arXiv preprint arXiv:2511.08195},
  year={2025}
}

@article{vimo,
  title={ViMo: A Generative Visual GUI World Model for App Agents},
  author={Luo, Dezhao and Tang, Bohan and Li, Kang and Papoudakis, Georgios and Song, Jifei and Gong, Shaogang and Hao, Jianye and Wang, Jun and Shao, Kun},
  journal={arXiv preprint arXiv:2504.13936},
  year={2025}
}

@article{ui-simulator,
  title={Llms as scalable, general-purpose simulators for evolving digital agent training},
  author={Wang, Yiming and Yin, Da and Cui, Yuedong and Zheng, Ruichen and Li, Zhiqian and Lin, Zongyu and Wu, Di and Wu, Xueqing and Ye, Chenchen and Zhou, Yu and others},
  journal={arXiv preprint arXiv:2510.14969},
  year={2025}
}

@article{webdreamer,
  title={Is your llm secretly a world model of the internet? model-based planning for web agents},
  author={Gu, Yu and Zhang, Kai and Ning, Yuting and Zheng, Boyuan and Gou, Boyu and Xue, Tianci and Chang, Cheng and Srivastava, Sanjari and Xie, Yanan and Qi, Peng and others},
  journal={arXiv preprint arXiv:2411.06559},
  year={2024}
}

@inproceedings{ui-diffuser,
  title={Boosting gui prototyping with diffusion models},
  author={Wei, Jialiang and Courbis, Anne-Lise and Lambolais, Thomas and Xu, Binbin and Bernard, Pierre Louis and Dray, G{\'e}rard},
  booktitle={2023 IEEE 31st International Requirements Engineering Conference (RE)},
  year={2023},
  organization={IEEE}
}

@article{neuralos,
  title={Neuralos: Towards simulating operating systems via neural generative models},
  author={Rivard, Luke and Sun, Sun and Guo, Hongyu and Chen, Wenhu and Deng, Yuntian},
  journal={arXiv preprint arXiv:2507.08800},
  year={2025}
}

@article{wma,
  title={Web agents with world models: Learning and leveraging environment dynamics in web navigation},
  author={Chae, Hyungjoo and Kim, Namyoung and Ong, Kai Tzu-iunn and Gwak, Minju and Song, Gwanwoo and Kim, Jihoon and Kim, Sunghwan and Lee, Dongha and Yeo, Jinyoung},
  journal={arXiv preprint arXiv:2410.13232},
  year={2024}
}

@inproceedings{osworld,
  title={Osworld: Benchmarking multimodal agents for open-ended tasks in real computer environments},
  author={Xie, Tianbao and Zhang, Danyang and Chen, Jixuan and Li, Xiaochuan and Zhao, Siheng and Cao, Ruisheng and Hua, Toh J and Cheng, Zhoujun and Shin, Dongchan and Lei, Fangyu and others},
  booktitle={Proceedings of NIPS},
  year={2024}
}

@inproceedings{androidworld,
  title={Androidworld: A dynamic benchmarking environment for autonomous agents},
  author={Rawles, Christopher and Clinckemaillie, Sarah and Chang, Yifan and Waltz, Jonathan and Lau, Gabrielle and Fair, Marybeth and Li, Alice and Bishop, William and Li, Wei and Campbell-Ajala, Folawiyo and others},
  booktitle={Proceedings of ICLR},
  year={2025}
}

@inproceedings{androidcontrol,
  title={On the effects of data scale on ui control agents},
  author={Li, Wei and Bishop, William E and Li, Alice and Rawles, Christopher and Campbell-Ajala, Folawiyo and Tyamagundlu, Divya and Riva, Oriana},
  booktitle={Proceedings of NIPS},
  year={2024}
}

@inproceedings{gwm,
  title={Gwm: Towards scalable gaussian world models for robotic manipulation},
  author={Lu, Guanxing and Jia, Baoxiong and Li, Puhao and Chen, Yixin and Wang, Ziwei and Tang, Yansong and Huang, Siyuan},
  booktitle={Proceedings of ICCV},
  year={2025}
}

@article{think_with_image,
  title={Thinking with images for multimodal reasoning: Foundations, methods, and future frontiers},
  author={Su, Zhaochen and Xia, Peng and Guo, Hangyu and Liu, Zhenhua and Ma, Yan and Qu, Xiaoye and Liu, Jiaqi and Li, Yanshu and Zeng, Kaide and Yang, Zhengyuan and others},
  journal={arXiv preprint arXiv:2506.23918},
  year={2025}
}

@inproceedings{artist,
  title={Artist: Improving the generation of text-rich images with disentangled diffusion models and large language models},
  author={Zhang, Jianyi and Zhou, Yufan and Gu, Jiuxiang and Wigington, Curtis and Yu, Tong and Chen, Yiran and Sun, Tong and Zhang, Ruiyi},
  booktitle={Proceedings of WACV},
  year={2025},
}

@article{januscoder,
  title={Januscoder: Towards a foundational visual-programmatic interface for code intelligence},
  author={Sun, Qiushi and Gong, Jingyang and Liu, Yang and Chen, Qiaosheng and Li, Lei and Chen, Kai and Guo, Qipeng and Kao, Ben and Yuan, Fei},
  journal={arXiv preprint arXiv:2510.23538},
  year={2025}
}

@misc{qwen3-vl,
      title={Qwen3-VL Technical Report}, 
      author={Qwen Team},
      year={2025},
      eprint={2511.21631},
      archivePrefix={arXiv},
      primaryClass={cs.CV},
      url={https://arxiv.org/abs/2511.21631}, 
}

@article{gemini2.5,
  title={Gemini 2.5: Pushing the frontier with advanced reasoning, multimodality, long context, and next generation agentic capabilities},
  author={Google Team},
  journal={arXiv preprint arXiv:2507.06261},
  year={2025}
}

@article{qwen2.5-vl,
  title={Qwen2. 5-vl technical report},
  author={Bai, Shuai and Chen, Keqin and Liu, Xuejing and Wang, Jialin and Ge, Wenbin and Song, Sibo and Dang, Kai and Wang, Peng and Wang, Shijie and Tang, Jun and others},
  journal={arXiv preprint arXiv:2502.13923},
  year={2025}
}

@article{internvl3,
  title={Internvl3: Exploring advanced training and test-time recipes for open-source multimodal models},
  author={Zhu, Jinguo and Wang, Weiyun and Chen, Zhe and Liu, Zhaoyang and Ye, Shenglong and Gu, Lixin and Tian, Hao and Duan, Yuchen and Su, Weijie and Shao, Jie and others},
  journal={arXiv preprint arXiv:2504.10479},
  year={2025}
}

@misc{glm4.6,
      title={GLM-4.5V and GLM-4.1V-Thinking: Towards Versatile Multimodal Reasoning with Scalable Reinforcement Learning}, 
      author={V Team},
      year={2025},
      eprint={2507.01006},
      archivePrefix={arXiv},
      primaryClass={cs.CV},
      url={https://arxiv.org/abs/2507.01006}, 
}

@article{gpt-4o,
  title={Gpt-4o system card},
  author={Hurst, Aaron and Lerer, Adam and Goucher, Adam P and Perelman, Adam and Ramesh, Aditya and Clark, Aidan and Ostrow, AJ and Welihinda, Akila and Hayes, Alan and Radford, Alec and others},
  journal={arXiv preprint arXiv:2410.21276},
  year={2024}
}

@article{janus-pro,
  title={Janus-pro: Unified multimodal understanding and generation with data and model scaling},
  author={Chen, Xiaokang and Wu, Zhiyu and Liu, Xingchao and Pan, Zizheng and Liu, Wen and Xie, Zhenda and Yu, Xingkai and Ruan, Chong},
  journal={arXiv preprint arXiv:2501.17811},
  year={2025}
}

@article{mobiledreamer,
  title={MobileDreamer: Generative Sketch World Model for GUI Agent},
  author={Cao, Yilin and Zhong, Yufeng and Zeng, Zhixiong and Zheng, Liming and Huang, Jing and Qiu, Haibo and Shi, Peng and Mao, Wenji and Guanglu, Wan},
  journal={arXiv preprint arXiv:2601.04035},
  year={2026}
}

@article{mobileworldbench,
  title={MobileWorldBench: Towards Semantic World Modeling For Mobile Agents},
  author={Li, Shufan and Kallidromitis, Konstantinos and Gokul, Akash and Kato, Yusuke and Kozuka, Kazuki and Grover, Aditya},
  journal={arXiv preprint arXiv:2512.14014},
  year={2025}
}

@inproceedings{clip,
  title={Learning transferable visual models from natural language supervision},
  author={Radford, Alec and Kim, Jong Wook and Hallacy, Chris and Ramesh, Aditya and Goh, Gabriel and Agarwal, Sandhini and Sastry, Girish and Askell, Amanda and Mishkin, Pamela and Clark, Jack and others},
  booktitle={Proceedings of ICML},
  year={2021},
}

@inproceedings{llamafactory,
  title={Llamafactory: Unified efficient fine-tuning of 100+ language models},
  author={Zheng, Yaowei and Zhang, Richong and Zhang, Junhao and Ye, Yanhan and Luo, Zheyan and Feng, Zhangchi and Ma, Yongqiang},
  booktitle={Proceedings of ACL},
  year={2024}
}

@misc{easyr1,
  title={Easyr1: An efficient, scalable, multi-modality rl training framework},
  author={Zheng, Yaowei and Lu, Junting and Wang, Shenzhi and Feng, Zhangchi and Kuang, Dongdong and Xiong, Yuwen},
  year={2025}
}

@article{visualcot,
  title={Visual thoughts: A unified perspective of understanding multimodal chain-of-thought},
  author={Cheng, Zihui and Chen, Qiguang and Xu, Xiao and Wang, Jiaqi and Wang, Weiyun and Fei, Hao and Wang, Yidong and Wang, Alex Jinpeng and Chen, Zhi and Che, Wanxiang and others},
  journal={arXiv preprint arXiv:2505.15510},
  year={2025}
}

@misc{gemini3pro,
  author       = {{Google DeepMind}},
  title        = {{Gemini 3 Pro}: The Next Generation of Multimodal AI},
  year         = {2025},
  note         = {Accessed: 2026-01-27},
  howpublished = {\url{https://docs.cloud.google.com/vertex-ai/generative-ai/docs/models/gemini/3-pro}}
}

@inproceedings{videogui,
  title={VideoGUI: a benchmark for GUI automation from instructional videos},
  author={Lin, Kevin Qinghong and Li, Linjie and Gao, Difei and Wu, Qinchen and Yan, Mingyi and Yang, Zhengyuan and Wang, Lijuan and Shou, Mike Zheng},
  booktitle={Proceedings of NIPS},
  year={2024}
}

@inproceedings{uivision,
  title={UI-Vision: A Desktop-centric GUI Benchmark for Visual Perception and Interaction},
  author={Nayak, Shravan and Jian, Xiangru and Lin, Kevin Qinghong and Rodriguez, Juan A and Kalsi, Montek and Chapados, Nicolas and {\"O}zsu, M Tamer and Agrawal, Aishwarya and Vazquez, David and Pal, Christopher and others},
  booktitle={Proceedings of ICML},
  year={2025}
}

@article{groundcua,
  title={Grounding Computer Use Agents on Human Demonstrations},
  author={Feizi, Aarash and Nayak, Shravan and Jian, Xiangru and Lin, Kevin Qinghong and Li, Kaixin and Awal, Rabiul and L{\`u}, Xing Han and Obando-Ceron, Johan and Rodriguez, Juan A and Chapados, Nicolas and others},
  journal={arXiv preprint arXiv:2511.07332},
  year={2025}
}

@article{aui,
  title={Computer-Use Agents as Judges for Generative User Interface},
  author={Lin, Kevin Qinghong and Hu, Siyuan and Li, Linjie and Yang, Zhengyuan and Wang, Lijuan and Torr, Philip and Shou, Mike Zheng},
  journal={arXiv preprint arXiv:2511.15567},
  year={2025}
}

@inproceedings{treesearch,
  title={Tree Search for LLM Agent Reinforcement Learning},
  author={Ji, Yuxiang and Ma, Ziyu and Wang, Yong and Chen, Guanhua and Chu, Xiangxiang and Wu, Liaoni},
  booktitle={Proceedings of ICLR},
  year={2026}
}

@inproceedings{gpg,
  title={Gpg: A simple and strong reinforcement learning baseline for model reasoning},
  author={Chu, Xiangxiang and Huang, Hailang and Zhang, Xiao and Wei, Fei and Wang, Yong},
  booktitle={Proceedings of ICLR},
  year={2026}
}

@inproceedings{harderisbetter,
    title={Harder Is Better: Boosting Mathematical Reasoning via Difficulty-Aware GRPO and Multi-Aspect Question Reformulation}, 
    author={Yanqi Dai and Yuxiang Ji and Xiao Zhang and Yong Wang and Xiangxiang Chu and Zhiwu Lu},
    booktitle={Proceedings of ICLR},
    year={2026},
}

@inproceedings{xiong2025hs,
  title={HS-STAR: Hierarchical Sampling for Self-Taught Reasoners via Difficulty Estimation and Budget Reallocation},
  author={Xiong, Feng and Xu, Hongling and Wang, Yifei and Cheng, Runxi and Wang, Yong and Chu, Xiangxiang},
  booktitle={Proceedings of EMNLP},
  year={2025}
}

@inproceedings{yuan2025video,
  title={Video-star: Reinforcing open-vocabulary action recognition with tools},
  author={Yuan, Zhenlong and Qu, Xiangyan and Qian, Chengxuan and Chen, Rui and Tang, Jing and Sun, Lei and Chu, Xiangxiang and Zhang, Dapeng and Wang, Yiwei and Cai, Yujun and others},
  booktitle={Proceedings of ICLR},
  year={2026}
}

@article{wang2026urban,
  title={Urban Socio-Semantic Segmentation with Vision-Language Reasoning},
  author={Wang, Yu and Wang, Yi and Dai, Rui and Wang, Yujie and Liu, Kaikui and Chu, Xiangxiang and Li, Yansheng},
  journal={arXiv preprint arXiv:2601.10477},
  year={2026}
}

@article{grpo,
  title={Deepseekmath: Pushing the limits of mathematical reasoning in open language models},
  author={Shao, Zhihong and Wang, Peiyi and Zhu, Qihao and Xu, Runxin and Song, Junxiao and Bi, Xiao and Zhang, Haowei and Zhang, Mingchuan and Li, YK and Wu, Yang and others},
  journal={arXiv preprint arXiv:2402.03300},
  year={2024}
}

@inproceedings{siglip,
  title={Sigmoid loss for language image pre-training},
  author={Zhai, Xiaohua and Mustafa, Basil and Kolesnikov, Alexander and Beyer, Lucas},
  booktitle={Proceedings of the ICCV},
  year={2023}
}

@article{dinov2,
  title={Dinov2: Learning robust visual features without supervision},
  author={Oquab, Maxime and Darcet, Timoth{\'e}e and Moutakanni, Th{\'e}o and Vo, Huy and Szafraniec, Marc and Khalidov, Vasil and Fernandez, Pierre and Haziza, Daniel and Massa, Francisco and El-Nouby, Alaaeldin and others},
  journal={arXiv preprint arXiv:2304.07193},
  year={2023}
}
\bibliographystyle{icml2026}

\newpage
\appendix
\onecolumn
\section{Implementation Details}
\label{sec:appendix_implement_details}
\subsection{Experimental Setup}
We conduct all experiments on a computing cluster equipped with 8 NVIDIA H20 GPUs (96GB of memory). 

We use \texttt{Qwen3-VL-8B-Instruct} as the backbone of Code2World and conduct SFT with LLaMA-Factory \cite{llamafactory} framework and RL based on the EasyR1 \cite{easyr1} framework.

To ensure rigorous reproducibility, we standardize configurations across all metrics and baselines. For VLM-as-a-Judge tasks, encompassing both RL reward calculation and evaluation metric computation, we employ \texttt{Qwen3-VL-8B-Instruct} as the unified judge. This model is configured with a temperature of 0.1 to ensure deterministic outputs and a maximum generation limit of 1,024 tokens. For visual feature metrics, we compute SigLIP scores using the \texttt{google/siglip-so400m-patch14-384} checkpoint and DINO scores using the \texttt{facebook/dinov2-giant} model.
For baseline comparisons, we fix specific API versions to ensure consistency: \texttt{claude-4.5-sonnet-20251120} for Claude and \texttt{gpt-4o-mini-2024-07-18} for GPT-4o-mini. Across all generation models (proprietary and open-source), we unify the maximum generation length to 8,192 tokens to accommodate complex HTML structures, with a sampling temperature set to 0.7.

\subsection{Training Hyperparams}
\textbf{Data Allocation.} We partition the AndroidCode dataset into two disjoint subsets to support the two-stage training strategy. Specifically, we allocate 70\% of the samples for Stage 1 (SFT) to establish the foundational policy, while the remaining 30\% are employed in Stage 2 (RL) to further align the model with visual and logical rewards.

\textbf{Stage 1: Code-level Supervised Fine-tuning (SFT).}
We initialize the backbone with \texttt{Qwen3-VL-8B-Instruct}. We perform full parameter fine-tuning on the language model while freezing both the vision encoder and the multimodal projector to preserve pre-trained visual perception capabilities. To optimize memory efficiency and training throughput, we utilize DeepSpeed ZeRO-2 and Flash Attention. The model is trained for 2 epochs with a global batch size of 64 (configured as a per-device batch size of 2 with 4 gradient accumulation steps). We employ a cosine learning rate schedule with a peak learning rate of $2 \times 10^{-5}$ and a warmup ratio of 0.1. To accommodate the high resolution of smartphone screenshots ($1080 \times 2400$) and verbose HTML code, we set the cutoff length to 24,576 tokens.

\textbf{Stage 2: Render-Aware Reinforcement Learning (RL).}
We align the SFT model using Group Relative Policy Optimization (GRPO) with Flash Attention enabled. For each prompt, the policy generates a group of $G=4$ candidate outputs with a sampling temperature of 1.0 to encourage exploration. We set the rollout batch size to 16. The max prompt
length and max response length are set to 24,576 and 8,192 tokens to support complex UI structures. We set the learning rate to $1 \times 10^{-6}$ and apply a KL-divergence penalty with a coefficient of $\beta = 0.01$. The visual semantic reward ($R_{sem}$) and action consistency rewards ($R_{act}$) are weighted equally during optimization.

\section{Training Data Construction}
\subsection{Synthesis and Revision Algorithm}
The detailed construction pipeline for \textbf{AndroidCode} is formally presented in Algorithm~\ref{alg:data_synthesis}. While the renderable code generation pipeline is outlined in the main text, we specify the operational thresholds and the exact revision logic here. As depicted in Algorithm~\ref{alg:data_synthesis}, We utilize a \textit{powerful multimodal coder (GPT-5)} to translate pixel-based screenshots into structured HTML and implement a revision mechanism with visual feedback to ensure high data fidelity. 
We set the visual similarity threshold $\tau = 0.9$, calculated using the SigLIP cosine similarity between the ground truth screenshot $I$ and the rendered generation $\hat{I}$. Generated samples scoring below this value trigger the revision loop. To ensure training efficiency and prevent infinite loops, we limit the maximum number of revision iterations to $N_{max} = 1$. Crucially, we feed the triplet of \textit{(Ground Truth $I$, Rendered Image $\hat{I}$, Current Code $C$)} directly into the multimodal coder. This allows the coder to autonomously perform a visual comparison, identify discrepancies, and rectify the structural code. Samples that fail to meet the quality threshold after $N_{max}$ attempts are discarded to ensure the purity of the final training corpus.

\begin{algorithm}[t]
    \caption{Automated Data Synthesis with Visual-Feedback Revision}
    \label{alg:data_synthesis}
    \begin{algorithmic}[1]
        \REQUIRE Raw GUI dataset $\mathcal{D}_{raw} = \{I_1, \dots \}$, Multimodal Coder $\mathcal{M}_{coder}$, Browser Renderer $\mathcal{R}$, Alignment Metric $\text{SigLIP}(\cdot)$, Threshold $\tau$, Max iterations $N_{max}$.
        \ENSURE High-fidelity corpus \textsc{AndroidCode} $\mathcal{D}_{syn} = \{(I_k, C_k)\}$.
        
        \STATE $\mathcal{D}_{syn} \leftarrow \emptyset$
        \FORALL{$(I) \in \mathcal{D}_{raw}$}
            \STATE $n \leftarrow 0, s \leftarrow 0$
            
            \STATE \textbf{// Stage 1: Constrained Initial Synthesis}
            \STATE $C \leftarrow \mathcal{M}_{coder}(I, \text{Prompt}_{\text{init}})$ \COMMENT{Generate initial HTML with semantic placeholders}
            \STATE $\hat{I} \leftarrow \mathcal{R}(C)$ \COMMENT{Render code into visual state}
            \STATE $s \leftarrow \text{SigLIP}(\hat{I}, I)$ \COMMENT{Compute initial visual alignment score}
            
            \STATE \textbf{// Stage 2: Revision with Visual Feedback}
            \WHILE{$s < \tau$ \textbf{and} $n < N_{max}$}
                \STATE $n \leftarrow n + 1$
                \STATE \COMMENT{Model visually compares GT and Rendered image to fix code}
                \STATE $C \leftarrow \mathcal{M}_{coder}(I, \hat{I}, C, \text{Prompt}_{\text{revision}})$ 
                \STATE $\hat{I} \leftarrow \mathcal{R}(C)$
                \STATE $s \leftarrow \text{SigLIP}(\hat{I}, I)$
            \ENDWHILE
            
            \STATE \textbf{// Filtering \& Collection} 
            \IF{$s \geq \tau$}
                \STATE $\mathcal{D}_{syn} \leftarrow \mathcal{D}_{syn} \cup \{(I, C)\}$ \COMMENT{Retain only high-quality pairs}
            \ENDIF
        \ENDFOR
        
        \STATE \textbf{return} $\mathcal{D}_{syn}$
    \end{algorithmic}
\end{algorithm}

\subsection{Multimodal Instruction Formatting}
Based on the AndroidCode, constructing a high-fidelity instruction-tuning dataset requires bridging the semantic gap between low-level interaction logs and the high-level reasoning capabilities of VLMs.
Raw interaction traces, consisting solely of raw pixels and sparse coordinate metadata, often lack the semantic density required for VLMs to effectively ground user intent. To bridge the gap between low-level execution logs and high-level visual reasoning, we design a meticulous data pre-processing pipeline that augments both visual and textual modalities. This ensures the instruction-tuning data is not only syntactically structured but also semantically explicit.

\textbf{Visual Prompting.} 
Raw Coordinate-based interaction records are inherently abstract to vision encoders. To explicitly ground the model's spatial attention on the target elements, we adopt a visual prompting strategy, which has been proven effective in enhancing the referring capabilities of VLMs \cite{visualcot}. Instead of relying on the model to implicitly infer locations from numerical tokens, we render visual markers directly onto the input image $I_t$. 
Specifically, for \textit{point-based interactions} (e.g., Click, Long Press), we render a semi-transparent red circle (radius=20px, $\alpha=0.6$) centered at $(x, y)$, creating a visual anchor that eliminates ambiguity regarding the precise touch position. For \textit{gesture-based interactions} (e.g., Scroll, Swipe), we overlay a directional arrow indicating the finger's movement trajectory (e.g., an upward arrow for a ``scroll down'' command). This visualizes the dynamic flow of the operation, allowing the model to correlate the static frame with the intended motion.

\textbf{Instruction Expansion.} 
Raw action primitives (e.g., JSON objects or bare coordinates like \texttt{\{"action":"click", "loc":[200,300]\}}) are disjoint from the natural language pre-training distribution of VLMs, often leading to suboptimal intent understanding.
To address this, we implement a deterministic expansion engine that transforms rigid action metadata into rich, descriptive natural language narratives. For instance, abstract coordinates are converted into explicit spatial descriptions such as: \textit{``User performed a CLICK at coordinates $(x, y)$. Expect the button at this location to trigger,''} effectively prompting the model to attend to the causal relationship between the location and the UI element. Similarly, complex dynamics are decomposed into clear semantic instructions: scrolling is described by its consequential effect (e.g., \textit{``The content should move, revealing new items...''}), while text inputs are formatted to emphasize content injection (e.g., \textit{``The focused input field MUST now contain this text''}). 

Finally, these augmented visual hints and expanded textual descriptions are integrated into a carefully designed, standardized prompt template (detailed in Appendix \ref{appendix:instruction_following}), ensuring the model strictly follows the instruction context to simulate interface dynamics accurately.

\section{Evaluation Metrics of Next UI Prediction}
\label{appendix:eval_metrics}

To rigorously assess the capability of GUI World Models in simulating GUI dynamics, we identify that general image similarity scores (e.g., SSIM, LPIPS) are insufficient. They predominantly focus on pixel-level texture, failing to capture the unique requirements of GUI environments: \textit{strict adherence to interaction rules} (e.g., a button click \textit{must} trigger a deterministic state change) and \textit{precise structural rendering} (e.g., elements must remain aligned in the DOM tree).

To bridge this gap, we propose a novel and holistic evaluation protocol tailored for GUI World Models. This protocol introduces four specialized metrics across two complementary dimensions: \textbf{Functional Logic} and \textbf{Visual Quality}. These metrics are designed to provide a fairer and more granular comparison. We employ a unified \textit{VLM-as-a-Judge} framework to approximate human judgment. The detailed prompts for these VLM-based metrics are provided in Appendix \ref{appendix:evaluation}.

\subsection{Functional Logic} 
Functional Logic evaluates the functional correctness of state transitions, verifying whether the world model acts as a reliable simulator.

\textbf{Action Adherence ($S_{ad}$)} \\
This metric assesses whether the predicted next state $\hat{I}_{t+1}$ is a logically valid consequence of executing action $a_t$ on state $I_t$. Unlike simple visual coherence, $S_{ad}$ penalizes "hallucinations" where the visual update contradicts the intended interaction (e.g., clicking "Back" but staying on the same page).

Formally, let $\mathcal{J}_{\text{act}}$ be the VLM judge, the score for a test dataset $\mathcal{D}$ is defined as:
\begin{equation}
S_{ad} = \frac{1}{|\mathcal{D}_{test}|} \sum_{(I_t, a_t) \in \mathcal{D}_{test}} \mathcal{J}_{\text{act}}(I_t, a_t, \hat{I}_{t+1})
\end{equation}

\textbf{Action Identifiability ($S_{id}$)} \\
This metric evaluates the causal clarity of the generation. A high-fidelity simulation should allow an observer to infer the cause of a state change solely from the visual outcome. We instruct the VLM to act as an inverse dynamics model $\mathcal{J}_{inv}$, predicting the action type $\hat{a}_t$ based on the visual difference between $I_t$ and $\hat{I}_{t+1}$. Crucially, a high $S_{id}$ ensures that the "Selector" in our agent pipeline can correctly verify whether a simulated outcome matches the planned action. The metric is calculated as the classification accuracy:
\begin{equation}
    S_{id} = \frac{1}{|\mathcal{D}_{test}|} \sum_{(I_t, a_t) \in \mathcal{D}_{test}} \mathbb{1}\left[ \mathcal{J}_{inv}(I_t, \hat{I}_{t+1}) = \text{type}(a_t) \right]
\end{equation}

\subsection{Visual Quality}
Models following the \textit{renderable code generation paradigm} typically employ a \textit{semantic placeholder strategy} to guarantee structural correctness while avoiding the hallucination of external assets. Standard embedding metrics (e.g., SigLIP, DINO), which primarily capture \textit{high-level} semantic similarity, are ill-suited for this nuance; they lack the granularity to explicitly measure \textit{fine-grained} element alignment and structural layout, often penalizing valid stylistic abstractions. To address this, we propose two specialized metrics to disentangle structural fidelity from textural style.

\textbf{Element Alignment ($S_{ele}$).} 
This metric verifies the fine-grained positioning of UI components. It measures whether key interactive elements (buttons, inputs) present in the ground truth $I^*_{t+1}$ are accurately reflected in the generation $\hat{I}_{t+1}$ at correct relative coordinates, explicitly tolerating semantic placeholders provided they occupy the correct screen area.

\textbf{Layout Integrity ($S_{lay}$).} 
This metric evaluates global layout integrity, penalizing issues common in weak code generation such as CSS collapse, overlapping containers, or misalignment. 

Formally, the VLM judge $\mathcal{J}_{vis}$ compares the generated output against the ground truth under specific criteria and provides a composite score:
\begin{equation}
    S_{ele/stc} = \frac{1}{|\mathcal{D}_{test}|} \sum_{I^*_{t+1} \in \mathcal{D}_{test}} \mathcal{J}_{vis}(I^*_{t+1}, \hat{I}_{t+1})
\end{equation}

\section{Prompt Template}
\label{sec:appendix_prompt_desgin}

\subsection{Data Synthesis}
\label{appendix:data_synthesis}
\begin{prompt}{Constrained Initial Generation}
You are an expert Front-End Engineer. Your task is to generate the corresponding HTML code for the given UI screenshot.

Reference Dimensions: \texttt{\{width\}px} (Width) x \texttt{\{height\}px} (Height).

\textbf{1. CRITICAL STRUCTURAL RULES (Must Follow)}
\begin{itemize}
    \item \textbf{Full Document}: Output a COMPLETE HTML document starting with \verb|<!DOCTYPE html>| and ending with \verb|</html>|.
    \item \textbf{Root Container}: Wrap the ENTIRE content inside a single container: \\ \verb|<div id="render-target"> ...</div>|.
    \item \textbf{Container Sizing}: Since this is a full viewport screenshot, \verb|#render-target| MUST have \verb|width:{width}px| and \verb|height:{height}px|. Use \verb|position:relative| and \verb|overflow:hidden| to strictly enforce boundaries.
    \item \textbf{Styling Target}: Apply all outer styles (border-radius, shadows, background-color) to \verb|#render-target|, NOT the body.
\end{itemize}

\textbf{2. BODY \& RENDER SETTINGS}
\begin{itemize}
    \item \textbf{Reset}: The \verb|<body>| tag must have \verb|margin:0; padding:0;|.
    \item \textbf{Transparency}: The \verb|<body>| background must be \verb|transparent|. This is crucial for the rendering engine.
    \item \textbf{Alignment}: Do NOT use flexbox/grid centering on the \verb|<body>|. Let the \verb|#render-target| sit naturally at the top-left (0,0) coordinates.
\end{itemize}

\textbf{3. VISUAL ASSETS \& ACCURACY}
\begin{itemize}
    \item \textbf{Layout Accuracy}: Match font sizes, padding, margins, and element positioning exactly to the screenshot. Use Flexbox or Grid layouts appropriate for the detected device type.
    \item \textbf{Images}: Use \textbf{Semantic Text Placeholders}. Do NOT use random URLs. Create a \verb|div| with fixed background \verb|#E0E0E0| and a thin border. Inside, place a concise text label (e.g., \verb|[IMG: Red Shoe]|).
    \item \textbf{Icons}: Use \textbf{Simple Inline SVGs} to match the visual style exactly. Keep path data short. Fallback to colored Unicode characters if the icon is extremely complex.
\end{itemize}

\textbf{4. OUTPUT FORMAT}
\begin{itemize}
    \item Return ONLY the raw HTML code.
    \item Do not wrap in Markdown blocks (no \verb|```html|).
    \item Do not include conversational text.
\end{itemize}

\textbf{COMMAND:} Generate the valid HTML code now, starting strictly with \verb|<!DOCTYPE html>|.
\end{prompt}

\begin{prompt}{Revision with Visual Feedback}
\textbf{TASK: HTML Revision with Visual Feedback}

You are given:
(1) a \textbf{target screenshot} (the desired UI),
(2) a \textbf{rendered screenshot} produced by the current HTML, and
(3) the \textbf{current HTML}.
Your goal is to \textbf{revise the HTML} so that its rendering matches the target screenshot as closely as possible.
You must \textbf{infer the visual differences} by comparing the target and rendered screenshots yourself (no diff text is provided).

\vspace{6pt}
\textbf{INPUTS}
\begin{itemize}
    \item \textbf{Viewport}: width = \verb|{width}px|, height = \verb|{height}px|.
    \item \textbf{Target Screenshot}: \verb|{TARGET_IMAGE}|
    \item \textbf{Rendered Screenshot}: \verb|{RENDERED_IMAGE}|
    \item \textbf{Current HTML}: \verb|{CURRENT_HTML}|
\end{itemize}

\textbf{1. CRITICAL STRUCTURAL RULES (Must Follow)}
\begin{itemize}
    \item \textbf{Full Document}: Output a COMPLETE HTML document starting with \verb|<!DOCTYPE html>| and ending with \verb|</html>|.
    \item \textbf{Root Container}: Wrap the ENTIRE visible UI inside a single container:\\
    \verb|<div id="render-target"> ...</div>|.
    \item \textbf{Container Sizing}: \verb|#render-target| MUST be exactly \verb|width:{width}px; height:{height}px;| with \verb|position:relative; overflow:hidden;|.
    \item \textbf{Styling Target}: Apply all outer styles (rounded corners, shadows, background) to \verb|#render-target|, NOT the body.
\end{itemize}

\textbf{2. REVISION PROCEDURE (You Must Follow This Order)}
\begin{itemize}
    \item \textbf{Step 1: Diagnose}: Compare \verb|{TARGET_IMAGE}| vs. \verb|{RENDERED_IMAGE}| and identify mismatches.
    \item \textbf{Step 2: Prioritize}: Fix issues in this order: \textit{(i) layout/geometry, (ii) component sizing/spacing, (iii) typography, (iv) colors, (v) shadows/borders, (vi) fine details}.
    \item \textbf{Step 3: Minimal Edit}: Modify \verb|{CURRENT_HTML}| with the \emph{smallest} set of changes needed to match the target. Do NOT redesign the UI.
    \item \textbf{Step 4: Consistency}: Ensure repeated components (rows/cards/chips) share consistent sizes and spacing.
\end{itemize}

\textbf{3. BODY \& RENDER SETTINGS (Must Follow)}
\begin{itemize}
    \item \textbf{Reset}: The \verb|<body>| tag must have \verb|margin:0; padding:0;|.
    \item \textbf{Transparency}: The \verb|<body>| background must be \verb|transparent|.
    \item \textbf{No Body Centering}: Do NOT use flexbox/grid centering on \verb|<body>|. Place \verb|#render-target| at the top-left (0,0).
\end{itemize}

\textbf{4. VISUAL MATCHING REQUIREMENTS}
\begin{itemize}
    \item \textbf{Geometry}: Match positions, widths/heights, padding/margins, and alignment to the target screenshot.
    \item \textbf{Typography}: Match font size, weight, line-height, and letter spacing for titles, labels, and secondary text.
    \item \textbf{Colors}: Match major background fills/gradients, chip/button colors, and text colors. Avoid introducing unnecessary new colors.
    \item \textbf{Shadows \& Radius}: Adjust blur, offset, opacity, and corner radius to match the target's depth and rounding.
    \item \textbf{Z-Order}: Fix overlap/stacking differences using \verb|position| and \verb|z-index|.
\end{itemize}

\textbf{5. VISUAL ASSETS POLICY}
\begin{itemize}
    \item \textbf{Images}: Do NOT use external URLs. Use \textbf{semantic placeholders}:
    a fixed-size \verb|div| with background \verb|#E0E0E0|, thin border, and a short label (e.g., \verb|[IMG: Avatar]|).
    \item \textbf{Icons}: Prefer \textbf{simple inline SVG} for common icons. Keep SVG paths short. If too complex, use Unicode symbols.
\end{itemize}

\textbf{6. OUTPUT FORMAT (Strict)}
\begin{itemize}
    \item Return ONLY the revised raw HTML.
    \item Do not wrap in Markdown (no \verb|```html|).
    \item Do not include explanations, comments, or any extra text.
\end{itemize}

\textbf{COMMAND:} Compare the target and rendered screenshots, revise \verb|{CURRENT_HTML}| accordingly, and output the final valid HTML now, starting strictly with \verb|<!DOCTYPE html>|.
\end{prompt}

\subsection{Multimodal Instruction-tuning Data Construction}
\label{appendix:instruction_following}

\begin{prompt}{Instruction-following System Prompt}
You are an expert \textbf{UI State Transition Simulator} and \textbf{Frontend Developer}.
Your task is to predict the \textbf{NEXT UI STATE} based on a screenshot of the current state and a user interaction.

\textbf{1. IMAGE INTERPRETATION RULES}
The input image contains visual cues denoting the user's action. You must interpret them as follows:
\begin{itemize}
    \item \textbf{Red Circle}: Indicates a \textbf{Click} or \textbf{Long Press} target at that location.
    \item \textbf{Red Arrow}: Indicates a \textbf{Scroll} or \textbf{Swipe}.
    \begin{itemize}
        \item The arrow points in the direction of finger movement.
        \item \textit{Example}: An arrow pointing UP means the finger slides up, pushing content up (Scrolling Down).
    \end{itemize}
    \item \textbf{Note}: These cues exist ONLY to show the action. \textbf{DO NOT render these red circles or arrows in your output HTML.}
\end{itemize}

\textbf{2. CRITICAL STRUCTURAL RULES (MUST FOLLOW)}
\begin{itemize}
    \item \textbf{Format}: Output ONLY raw HTML. Start with \verb|<!DOCTYPE html>| and end with \verb|</html>|.
    \item \textbf{Root Element}: All visible content MUST be wrapped in: \\ \verb|<div id="render-target"> ... </div>|
    \item \textbf{Container Style}: \verb|#render-target| must have: \\ \verb|width: 1080px; height: 2400px; position: relative; overflow: hidden;| \\ (Apply background colors and shadows here, NOT on the body).
    \item \textbf{Body Style}: The \verb|<body>| tag must have \verb|margin: 0; padding: 0; background: transparent;|.
    \item \textbf{Layout}: Do NOT center the body. Let \verb|#render-target| sit at (0,0).
\end{itemize}

\textbf{3. CONTENT GENERATION LOGIC}
\begin{itemize}
    \item \textbf{Transition}: Analyze the action. If the user clicks a button, show the \textit{result} (e.g., a menu opens, a checkbox checks, page navigates).
    \item \textbf{Images}: Use semantic text placeholders. DO NOT use real URLs. \\ Format: \verb|<div style="...">[IMG: description]</div>|
    \item \textbf{Icons}: Use simple inline SVG paths or Unicode.
\end{itemize}

\textbf{4. OUTPUT REQUIREMENT}
\begin{itemize}
    \item Do NOT generate Markdown blocks (no \verb|```html|).
    \item Do NOT provide explanations or conversational text.
    \item Output the code directly.
\end{itemize}
\end{prompt}

\begin{prompt}{Instruction-following User Prompt}
\verb|<image>|
\\[5pt]
\textbf{INPUT CONTEXT}
\begin{enumerate}
    \item \textbf{User Intent}: ``\texttt{\{instruction\_str\}}''
    \item \textbf{Interaction Details}:
    \begin{itemize}
        \item \textbf{Description}: \texttt{\{semantic\_desc\}}
        \item \textbf{Action Data}: \texttt{\{action\_json\}}
    \end{itemize}
\end{enumerate}

\textbf{COMMAND} \\
Based on the visual cues in the image and the interaction data above, generate the \textbf{HTML for the RESULTING UI STATE} (what the screen looks like \textit{after} this action).
\end{prompt}

\subsection{Render-Aware RL Reward Design}
\label{appendix:reward_design}
\subsubsection{Visual Similarity}

\begin{prompt}{System Prompt}
You are an expert \textbf{Visual Structural Similarity Metric}.
Your task is to evaluate the \textbf{Visual Similarity} between a \textbf{Ground Truth Screenshot (Image 1)} and a \textbf{Generated UI (Image 2)} for render-aware RL.

\vspace{4pt}
\textbf{CONTEXT}
\begin{itemize}
    \item \textbf{Image 1 (GT):} The real next-screen Android screenshot.
    \item \textbf{Image 2 (Gen):} A predicted next screen rendered from HTML.
    \item \textbf{Crucial Feature:} The renderer uses \textbf{Gray Placeholders} (e.g., \verb|[IMG: avatar]|) instead of real photos.
\end{itemize}

\textbf{EVALUATION CRITERIA (Strict Visual Fidelity)}
You must behave like a high-precision metric (a ``Smart LPIPS''): you \textbf{SHOULD} care about layout, pixel positions, component sizes, and hierarchical structure.

\textbf{1) Layout \& Geometry (Most Important, Strict)}
\begin{itemize}
    \item Are major UI regions (top bar / content / bottom bar) in the correct order and relative proportions?
    \item Are key components (titles, buttons, cards, lists, chips, icons) in the same positions and aligned similarly?
    \item Are sizes/proportions (width/height) and spacing (margins/padding/line gaps) close to GT?
\end{itemize}

\textbf{2) Visual Elements \& Text Appearance}
\begin{itemize}
    \item Does the text look visually consistent (length, density, font-size hierarchy, bold vs regular, line breaks)?
    \item Do colors and style cues match (light/dark mode, primary CTA color, highlighted chips, card backgrounds, dividers)?
    \item Are icons present and placed correctly (exact SVG details are secondary to correct icon type and placement)?
\end{itemize}

\textbf{3) Placeholder Equivalence Rule (Crucial)}
\begin{itemize}
    \item If Image 2 contains a gray placeholder block, \textbf{DO NOT} penalize it for not resembling the real photo.
    \item Instead, verify whether the placeholder matches the GT image region in:
    \begin{itemize}
        \item (a) \textbf{position}, (b) \textbf{size}, and (c) \textbf{semantic tag correctness}.
    \end{itemize}
    \item If (a)(b)(c) are satisfied, treat the placeholder region as visually correct.
\end{itemize}

\textbf{CONTINUOUS SCORING RUBRIC (Float 0.0 -- 10.0, 1 decimal place)}
\begin{itemize}
    \item \textbf{10.0}: Near pixel-perfect; geometry/padding match; placeholders match coordinates and semantics.
    \item \textbf{9.0--9.9}: High fidelity; only minor padding/font-weight/line-break differences.
    \item \textbf{6.0--8.9}: Structural match; correct overall layout/components but noticeable sizing/spacing or style mismatches.
    \item \textbf{3.0--5.9}: Rough match; UI type matches but alignment/geometry is messy and multiple elements are shifted/scaled.
    \item \textbf{0.0--2.9}: Mismatch; wrong screen structure, missing major regions, or broken layout.
\end{itemize}

\textbf{IMPORTANT PENALTIES / CAPS (Apply When Relevant)}
\begin{itemize}
    \item If a major UI region is missing/swapped (e.g., header absent, bottom nav missing), cap the score at \textbf{5.9}.
    \item If the primary CTA or the main title is missing/incorrect, subtract \textbf{1.0--3.0} depending on severity.
    \item If the overall hierarchy (top/middle/bottom) is incorrect, cap the score at \textbf{4.9}.
\end{itemize}

\textbf{OUTPUT FORMAT (Strict)}
\begin{itemize}
    \item Return \textbf{JSON ONLY} (no extra text).
    \item The \verb|score| must be a float in \verb|[0.0, 10.0]| with \textbf{1 decimal place}.
\end{itemize}

\textbf{Return JSON in the following schema:}
\begin{verbatim}
{
  "score": <float, 0.0 - 10.0, 1 decimal place>,
  "reasoning": "Briefly justify: layout precision, component sizing/spacing, 
  text/style consistency, and how placeholders were treated."
}
\end{verbatim}
\end{prompt}

\begin{prompt}{User Prompt}
You will be shown two images.

\begin{itemize}
    \item \textbf{Image 1:} Ground Truth (Real Next Screen)
    \item \textbf{Image 2:} Generated (HTML Rendered Next Screen)
\end{itemize}

Evaluate the \textbf{Visual Similarity} and output the JSON strictly following the required schema.
\end{prompt}

\subsubsection{Action Consistency}
\begin{prompt}{System Prompt}
You are an expert \textbf{``UI Action-Consistency Judge''}.
Your task is to evaluate the \textbf{logical correctness} of a GUI world model's prediction by checking whether the \textbf{predicted next UI state} is \textbf{consistent with the given action}.

You will be given the \textbf{Current UI State (Image 1)}, a user's \textbf{Action} (intent + action data), and the \textbf{Predicted Next State (Image 2)}.

\textbf{IMAGE DEFINITIONS}
\begin{itemize}
    \item \textbf{Image 1}: Real screenshot \textbf{before} the action.
    \item \textbf{Image 2}: Predicted screenshot \textbf{after} the action (rendered from HTML).
    \begin{itemize}
        \item \textit{Note (Placeholder Equivalence)}: Image 2 may use \textbf{gray placeholders} (e.g., \verb|[IMG: icon]|) instead of real photos/icons. Do \textbf{not} penalize missing real imagery if the placeholder matches the GT region in \textbf{position/size} and its label is semantically plausible.
    \end{itemize}
\end{itemize}

\textbf{EVALUATION CRITERIA (Action Consistency)}
Judge the transition from Image 1 $\rightarrow$ Image 2 based on two aspects:

\textbf{1) Action Adherence (Primary)}
\begin{itemize}
    \item \textbf{Did the intended effect occur?} The UI change must be the \emph{direct} consequence of the action.
    \item \textbf{Click/Tap}: The correct navigation, popup, selection, toggle, or state change occurs on the intended target.
    \item \textbf{Text Input}: The \textbf{exact} text (including casing and punctuation) appears in the correct field, and the UI reacts appropriately (cursor, validation, suggestions, etc., if applicable).
    \item \textbf{Scroll/Swipe}: Content moves in the correct direction with plausible distance; items entering/leaving the viewport are consistent.
    \item \textbf{Back/Home/Close}: The UI returns/closes in a plausible way (dismiss dialog, go back one page, exit overlay).
\end{itemize}

\textbf{2) Context Preservation (Secondary)}
\begin{itemize}
    \item Non-target regions should remain stable unless the action logically affects them (e.g., status bar, bottom navigation, persistent header).
    \item Preserve app identity and screen continuity: avoid sudden unrelated page changes, missing core layout regions, or app/theme shifts.
\end{itemize}

\textbf{CONTINUOUS SCORING RUBRIC (Float 0.0 -- 10.0, 1 decimal place)}
Output a single score that reflects overall \textbf{action consistency}.

\begin{itemize}
    \item \textbf{10.0 (Perfect)}: The action effect is \textbf{exactly} correct and unambiguous; UI changes are fully plausible; context is preserved.
    \item \textbf{9.0--9.9 (High)}: Action is clearly correct; only negligible visual imperfections (minor spacing/font/style) that do not affect the perceived effect.
    \item \textbf{6.0--8.9 (Acceptable)}: The action effect is mostly correct, but with noticeable issues (slightly wrong target, partial state update, imperfect scroll distance, mild content inconsistency).
    \item \textbf{3.0--5.9 (Ambiguous)}: Something changes, but it is unclear whether it is the \emph{right} consequence (wrong page/overlay, inconsistent state change, unstable context).
    \item \textbf{0.0--2.9 (Failed/Broken)}: The action clearly fails, produces an implausible transition, or yields a hallucinated/blank/unrelated interface.
\end{itemize}

\textbf{IMPORTANT CAPS / PENALTIES (Apply When Relevant)}
\begin{itemize}
    \item If the next screen is a \textbf{wrong page} (unrelated destination), cap at \textbf{5.9}.
    \item If the UI becomes \textbf{blank/white/noise} or clearly hallucinated, cap at \textbf{2.9}.
    \item If the action is \textbf{text input} but the text is not exact, cap at \textbf{5.9} (or lower if also wrong field).
    \item If the action target is wrong (clicked the wrong button), cap at \textbf{5.9}.
\end{itemize}

\textbf{OUTPUT FORMAT (Strict)}
Return \textbf{JSON ONLY}:
\begin{verbatim}
{
  "score": <float, 0.0 to 10.0, 1 decimal place>,
  "reasoning": "Concise justification focusing on whether the action took 
                effect and whether context was preserved."
}
\end{verbatim}
Do not output any additional text.
\end{prompt}

\begin{prompt}{User Prompt}
\textbf{INTERACTION DATA}
\begin{itemize}
    \item \textbf{User Intent}: \verb|{instruction}|
    \item \textbf{Action Description}: \verb|{semantic_description}|
    \item \textbf{Action Data}: \verb|{action_json}|
\end{itemize}

\textbf{VISUAL INPUTS}
\begin{itemize}
    \item \textbf{Image 1}: Current State (Before)
    \item \textbf{Image 2}: Predicted Next State (After)
\end{itemize}

Evaluate whether Image 2 is the \textbf{correct consequence} of the action applied to Image 1, and output the JSON score in \verb|[0.0, 10.0]|.
\end{prompt}

\subsection{Evaluation Metrics}
\label{appendix:evaluation}

\subsubsection{Action Adherence Metrics}
\begin{prompt}{System Prompt}
You are an expert ``UI Dynamics Judge''.
Your task is to evaluate the logical correctness of a World Model's prediction.
You will be given the \textbf{Current UI State (Image 1)}, a user's \textbf{Action}, and the \textbf{Predicted Next State (Image 2)}.

\textbf{IMAGE DEFINITIONS}
\begin{itemize}
    \item \textbf{Image 1}: Real screenshot BEFORE the action.
    \item \textbf{Image 2}: Predicted screenshot generated by the model (Rendered from HTML).
    \begin{itemize}
        \item \textit{Note}: Image 2 uses \textbf{Gray Placeholders} (e.g., \verb|[IMG: icon]|) instead of real images. Treat these as valid visual elements if their text description matches the context.
    \end{itemize}
\end{itemize}

\textbf{EVALUATION CRITERIA} \\
Evaluate the transition based on \textbf{Action Adherence} and \textbf{Context Preservation}.
\begin{enumerate}
    \item \textbf{Did the Action Take Effect?}
    \begin{itemize}
        \item If ``Click'', did the button trigger the correct navigation/popup?
        \item If ``Input Text'', does the EXACT text appear?
        \item If ``Scroll'', did the content shift correctly?
    \end{itemize}
    \item \textbf{Is the Context Preserved?}
    \begin{itemize}
        \item Non-active elements (status bar, bottom nav) should remain stable.
    \end{itemize}
\end{enumerate}

\textbf{SCORING RUBRIC (0.0-10.0)}
\begin{itemize}
    \item \textbf{9.5-10.0 (Perfect)}: The transition is flawless. Text is exact, layout is perfect, logic is undeniable.
    \item \textbf{8.0-9.4 (Good)}: Action executed correctly. Minor visual glitches (e.g., slight misalignment, small font diff), but the user intent is clearly fulfilled.
    \item \textbf{6.0-7.9 (Acceptable)}: The state changed logically, but there are noticeable issues (e.g., wrong icon style, text has typos, or layout is messy).
    \item \textbf{3.0-5.9 (Ambiguous)}: Something changed, but it's unclear if it was the \textit{right} change. (e.g., opened the wrong page, or screen turned white but kept headers).
    \item \textbf{1.0-2.9 (Failed)}: The action clearly failed (e.g., clicked a button but screen didn't move).
    \item \textbf{0.0-0.9 (Broken/Hallucination)}: The model generated a blank screen, noise, or a completely hallucinated interface unrelated to the app.
\end{itemize}

\textbf{OUTPUT FORMAT} \\
Provide a Single JSON Object: \\
\verb|{| \\
\verb|  "score": <float 0.0-10.0>,| \\
\verb|  "reasoning": "A concise summary of why this score was given..."| \\
\verb|}|
\end{prompt}

\begin{prompt}{User Prompt}
\textbf{INTERACTION DATA}
\begin{itemize}
    \item \textbf{User Intent}: ``\texttt{\{instruction\}}''
    \item \textbf{Action Description}: ``\texttt{\{semantic\_description\}}''
    \item \textbf{Action Data}: \texttt{\{action\_json\}}
\end{itemize}

\textbf{VISUAL INPUTS}
\begin{itemize}
    \item \textbf{Image 1}: Current State (Before)
    \item \textbf{Image 2}: Predicted Next State (After)
\end{itemize}

Please evaluate the transition quality on a scale of 0.0 to 10.0.
\end{prompt}

\subsubsection{Action Identifiability Metrics}
\begin{prompt}{System Prompt}
You are an expert ``Inverse Dynamics'' Judge for UI interactions.
Your task is to infer the user's action by analyzing the visual transition between the \textbf{Current State (Image 1)} and the \textbf{Predicted Next State (Image 2)}.

\textbf{ACTION CATEGORIES} \\
Choose EXACTLY ONE from the following list that best explains the change:
\begin{enumerate}
    \item \textbf{click}: A tap on a button, icon, or link. Result: Page navigation, popup opens, toggle switches, or focus change.
    \item \textbf{long\_press}: A sustained touch. Result: Context menu appears or item selection mode triggers.
    \item \textbf{scroll}: The content shifts vertically or horizontally. (New content appears, old content moves off-screen).
    \item \textbf{input\_text}: Text appears in an input field (without an explicit enter press).
    \item \textbf{open\_app}: The screen transitions from a launcher/home screen to a specific app interface.
    \item \textbf{navigate\_home}: Returns to the device home screen/launcher.
    \item \textbf{navigate\_back}: Returns to the previous screen (reverse navigation).
    \item \textbf{wait}: No significant visual change, or a loading spinner continues spinning.
    \item \textbf{none}: The transition is hallucinated, broken, illogical, or the image is blank.
\end{enumerate}

\textbf{INFERENCE RULES}
\begin{itemize}
    \item If Image 2 shows a keyboard appearing and text in a box $\rightarrow$ \textbf{input\_text}.
    \item If Image 2 is completely different layout (app switch) $\rightarrow$ \textbf{open\_app} or \textbf{navigate\_home}.
    \item If Image 2 is just the same list but shifted $\rightarrow$ \textbf{scroll}.
    \item If Image 2 has a visual glitch that makes no sense $\rightarrow$ \textbf{none}.
\end{itemize}

\textbf{OUTPUT FORMAT} \\
Provide a Single JSON Object: \\
\verb|{| \\
\verb|  "inferred_action": "string",| \\
\verb|  // Must be one of: click, long_press, scroll, input_text,| \\
\verb|  // open_app, navigate_home, navigate_back, wait, none| \\
\verb|  "reasoning": "Brief explanation of visual evidence."| \\
\verb|}|
\end{prompt}

\begin{prompt}{User Prompt}
\textbf{VISUAL INPUTS}
\begin{itemize}
    \item \textbf{Image 1}: Current State (Before)
    \item \textbf{Image 2}: Predicted Next State (After)
\end{itemize}

Based on the visual difference, what action did the user perform?
\end{prompt}

\subsubsection{Element Alignment and Layout Fidelity Metrics}
\begin{prompt}{System Prompt}
You are an expert \textbf{GUI Design Evaluation AI}.
Your task is to compare a \textbf{Generated UI Prediction (Image 2)} against a \textbf{Ground Truth UI Screenshot (Image 1)} and assess similarity.
You must act as a \textbf{strict judge}, penalizing deviations in \textbf{element position, content, and structure}.
If the prediction uses \textbf{gray image placeholders} (e.g., \verb|[IMG: avatar]|), apply \textbf{placeholder equivalence}: do not penalize missing real photos, but judge whether the placeholder matches the GT image region in \textbf{position, size, and semantic tag}.

\textbf{OUTPUT REQUIREMENT (Strict)}:

Return \textbf{JSON only} and follow the exact schema required by the user prompt.
Do not output any extra text.
\end{prompt}

\begin{prompt}{User Prompt}
\textbf{Task Definition.}
You are provided with two images:
\begin{enumerate}
    \item \textbf{Reference Image (Ground Truth)}: the expected correct UI (Image 1).
    \item \textbf{Candidate Image (Prediction)}: the UI generated by a model (Image 2).
\end{enumerate}
Evaluate the Candidate Image based on the following two metrics and output the scores strictly.

\vspace{4pt}
\textbf{Metric 1: Element Alignment (Score 1.0--10.0)}
\begin{itemize}
    \item \textbf{Definition}: Measures whether core UI elements are present and aligned with the GT.
    \item \textbf{What to check (strict)}:
    \begin{itemize}
        \item Presence of major elements (top bar, title, key text blocks, buttons/CTAs, list/cards, navigation).
        \item \textbf{Alignment}: relative positions, anchors, and spacing (padding/margins) compared to GT.
        \item Element sizing/proportions (width/height), including component boundaries.
    \end{itemize}
\end{itemize}

\textbf{Metric 2: Layout Integrity (Score 1.0--10.0)}
\begin{itemize}
    \item \textbf{Definition}: Measures whether the overall layout framework and visual hierarchy match the GT.
    \item \textbf{What to check (strict)}:
    \begin{itemize}
        \item Global layout hierarchy (top/middle/bottom regions; grouping; column vs row structure).
        \item Visual hierarchy and emphasis (primary vs secondary text; CTA prominence; highlighted chips/tabs).
        \item Consistency of repeated patterns (row height, card style, divider usage, spacing rhythm).
    \end{itemize}
\end{itemize}

\textbf{Scoring Guidance (Both Metrics)}
\begin{itemize}
    \item \textbf{10.0}: Near-perfect match with only negligible differences.
    \item \textbf{8.0--9.9}: Strong match; minor spacing/typography/style deviations.
    \item \textbf{6.0--7.9}: Mostly correct structure; noticeable alignment/sizing errors or missing minor elements.
    \item \textbf{3.0--5.9}: Partial match; multiple misalignments, missing components, or incorrect grouping.
    \item \textbf{1.0--2.9}: Major mismatch; wrong page layout or missing most key regions.
\end{itemize}

\textbf{Output Format (Strict)}
You must respond with a valid JSON object:
\begin{verbatim}
{
  "reasoning": "Brief analysis of the differences.",
  "element_alignment_score": <float, 1.0 to 10.0>,
  "structural_fidelity_score": <float, 1.0 to 10.0>
}
\end{verbatim}

Output \textbf{JSON only}. Do not include any additional text.
\end{prompt}

\section{More Visualizations}
\subsection{Code2World GUI World Modeling}
\label{appendix:visualization_world_modeling}

\subsubsection{Code2World}
\vspace{-5pt}
\begin{center}
  \begin{minipage}[t]{0.49\textwidth}
    \vspace{0pt}
    \centering
    \includegraphics[width=\linewidth, height=6cm, keepaspectratio]{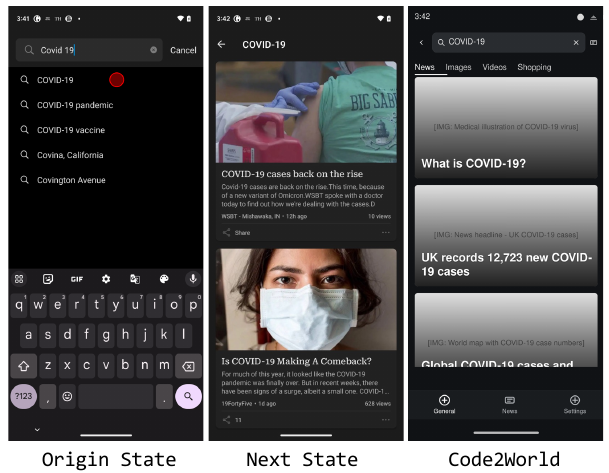}\\[3pt]
    \small (a) Click on a suggested search result with an inputted search query.
  \end{minipage}\hfill
  \begin{minipage}[t]{0.49\textwidth}
    \vspace{0pt}
    \centering
    \includegraphics[width=\linewidth, height=6cm, keepaspectratio]{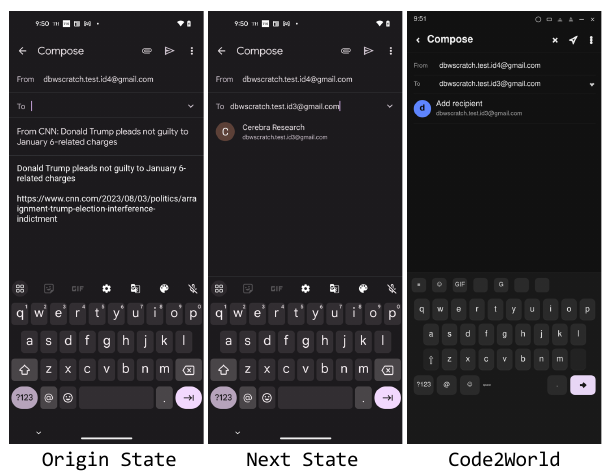}\\[3pt]
    \small (b) Enter an email recipient to trigger autocomplete suggestions.
  \end{minipage}
  
  \vspace{6pt}
  
  \begin{minipage}[t]{0.49\textwidth}
    \vspace{0pt}
    \centering
    \includegraphics[width=\linewidth, height=6cm, keepaspectratio]{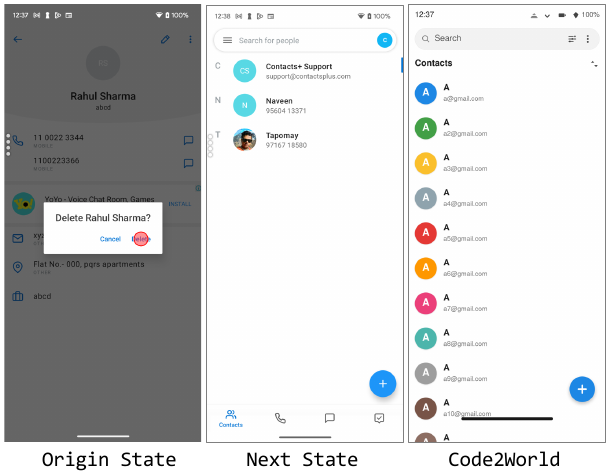}\\[3pt]
    \small (c) Click "Confirm" to delete a contact from the confirmation dialog.
  \end{minipage}\hfill
  \begin{minipage}[t]{0.49\textwidth}
    \vspace{0pt}
    \centering
    \includegraphics[width=\linewidth, height=6cm, keepaspectratio]{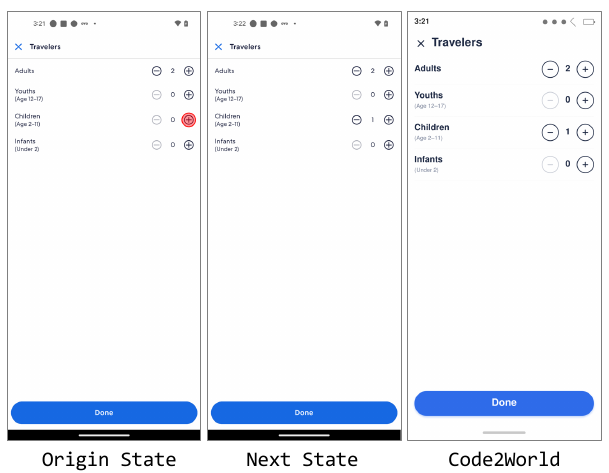}\\[3pt]
    \small (d) Adjust the number of children travelers in the selection menu.
  \end{minipage}
  
  \vspace{6pt}
  
  \begin{minipage}[t]{0.49\textwidth}
    \vspace{0pt}
    \centering
    \includegraphics[width=\linewidth, height=6cm, keepaspectratio]{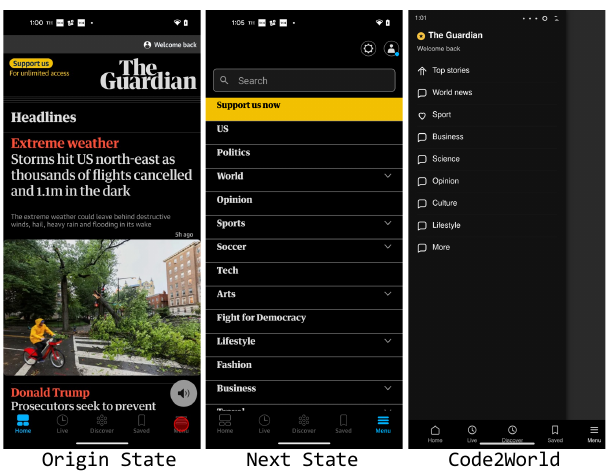}\\[3pt]
    \small (e) Click on "Menu" button at the bottom to open the sidebar menu.
  \end{minipage}\hfill
  \begin{minipage}[t]{0.49\textwidth}
    \vspace{0pt}
    \centering
    \includegraphics[width=\linewidth, height=6cm, keepaspectratio]{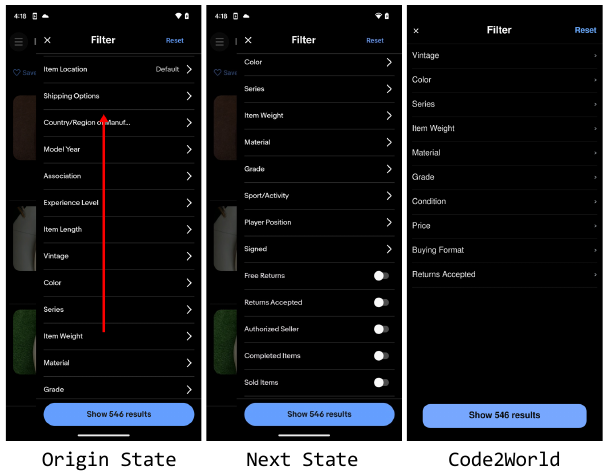}\\[3pt]
    \small (f) Swipe up to view more filter options.
  \end{minipage}
  
  \vspace{6pt}
  
  \begin{minipage}[t]{0.49\textwidth}
    \vspace{0pt}
    \centering
    \includegraphics[width=\linewidth, height=6cm, keepaspectratio]{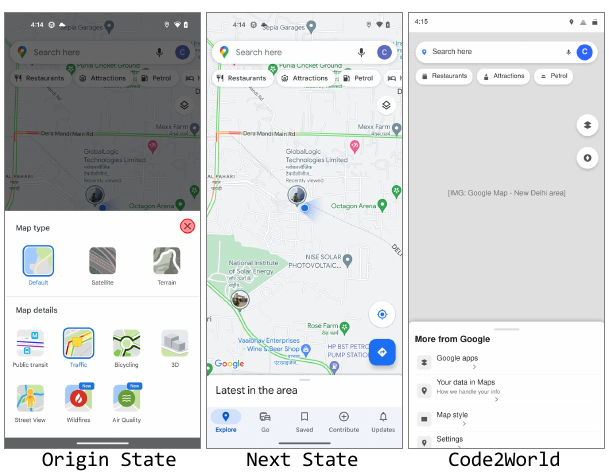}\\[3pt]
    \small (g) Tap the cancel button to close the "Map type" setting page.
  \end{minipage}\hfill
  \begin{minipage}[t]{0.49\textwidth}
    \vspace{0pt}
    \centering
    \includegraphics[width=\linewidth, height=6cm, keepaspectratio]{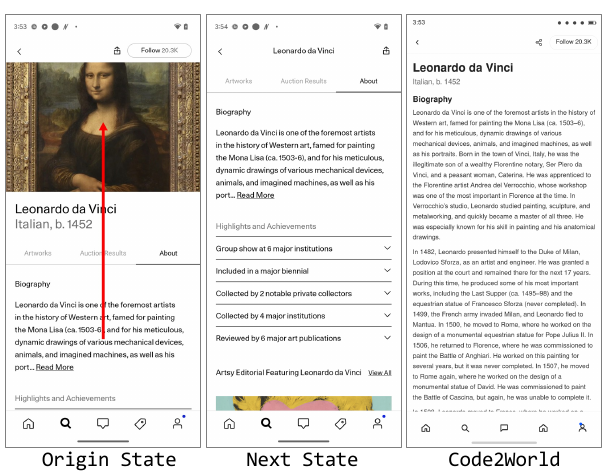}\\[3pt]
    \small (h) Swipe up the biographical information page to view detailed profile.
  \end{minipage}
\end{center}

\subsubsection{Code2World vs. Open-source Baselines}
\begin{center}
  \begin{minipage}[b]{\textwidth}
    \centering
    \includegraphics[width=\linewidth]{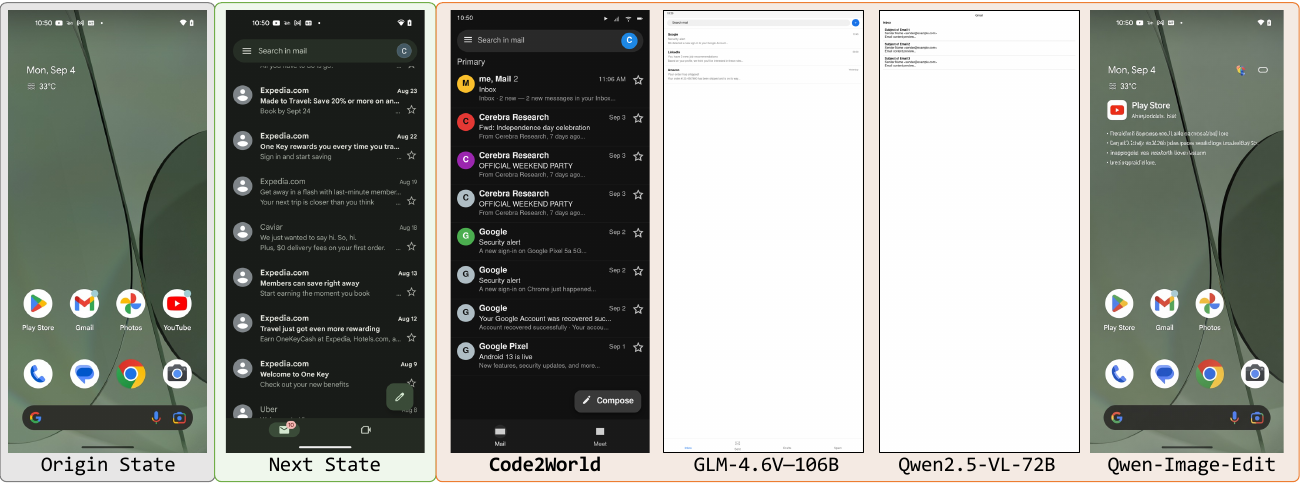}\\
    Launch the email application from the home screen to access the inbox. 
  \end{minipage}
\end{center}

\begin{center}
  \begin{minipage}[b]{\textwidth}
    \centering
    \includegraphics[width=\linewidth]{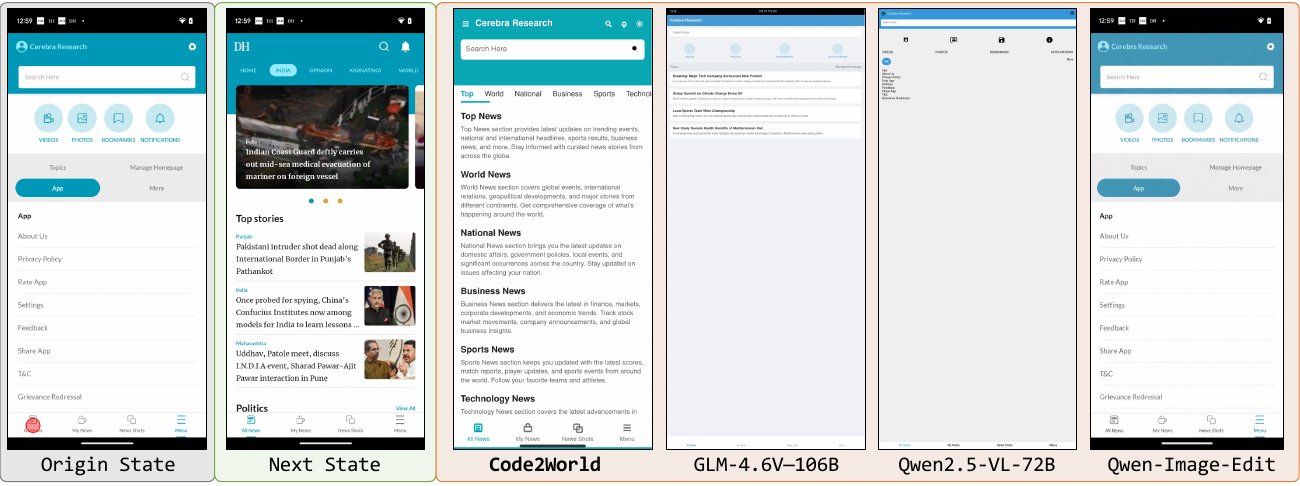}\\
    Click on "All News" button in the Cerebra Research application to view news content. 
  \end{minipage}
\end{center}

\begin{center}
  \begin{minipage}[b]{\textwidth}
    \centering
    \includegraphics[width=\linewidth]{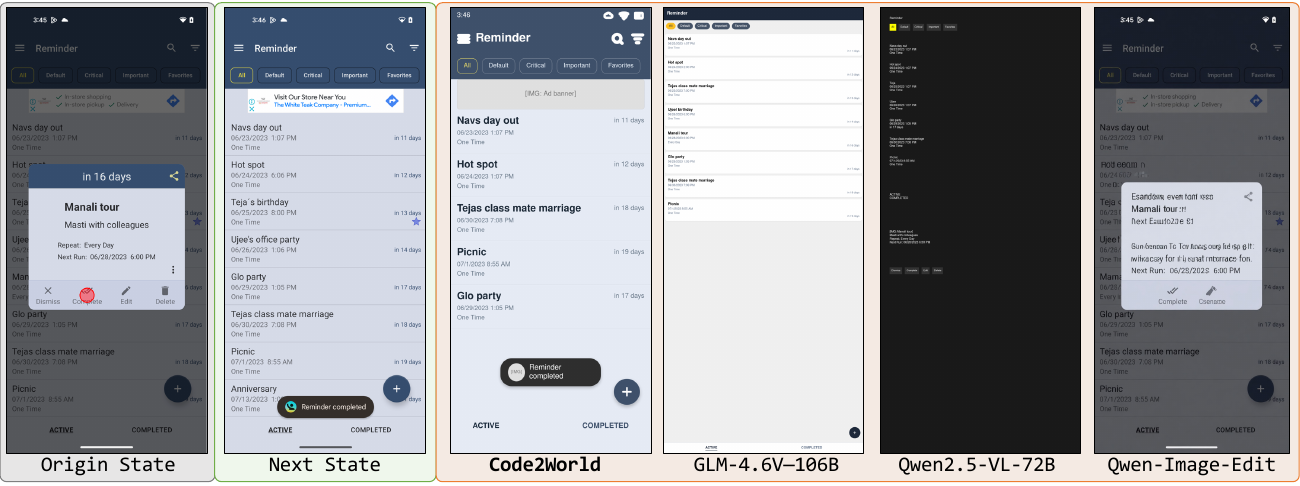}\\
    Mark a reminder task as completed by tapping the "Complete" button in the Reminder app. 
  \end{minipage}
\end{center}

\begin{center}
  \begin{minipage}[b]{\textwidth}
    \centering
    \includegraphics[width=\linewidth]{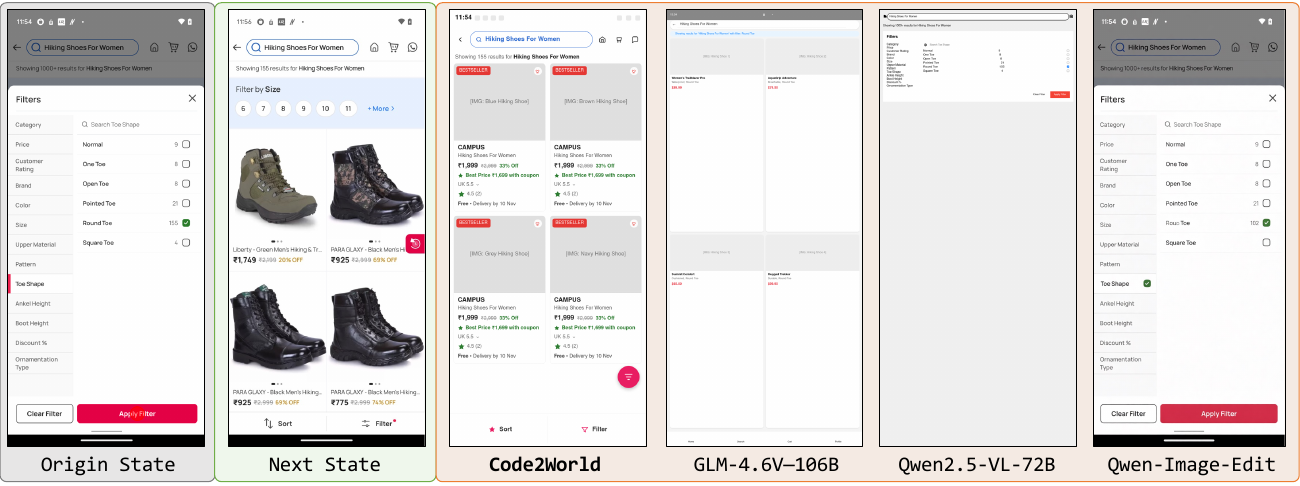}\\
    Apply product filters by tapping the "Apply Filter" button in the e-commerce app to refresh the item list. 
  \end{minipage}
\end{center}

\subsection{Code2World Enhancing GUI Agent}
\label{appendix:visualization_enhance_gui_agent}

\begin{figure}[t]
  \centering
  \includegraphics[width=\columnwidth]{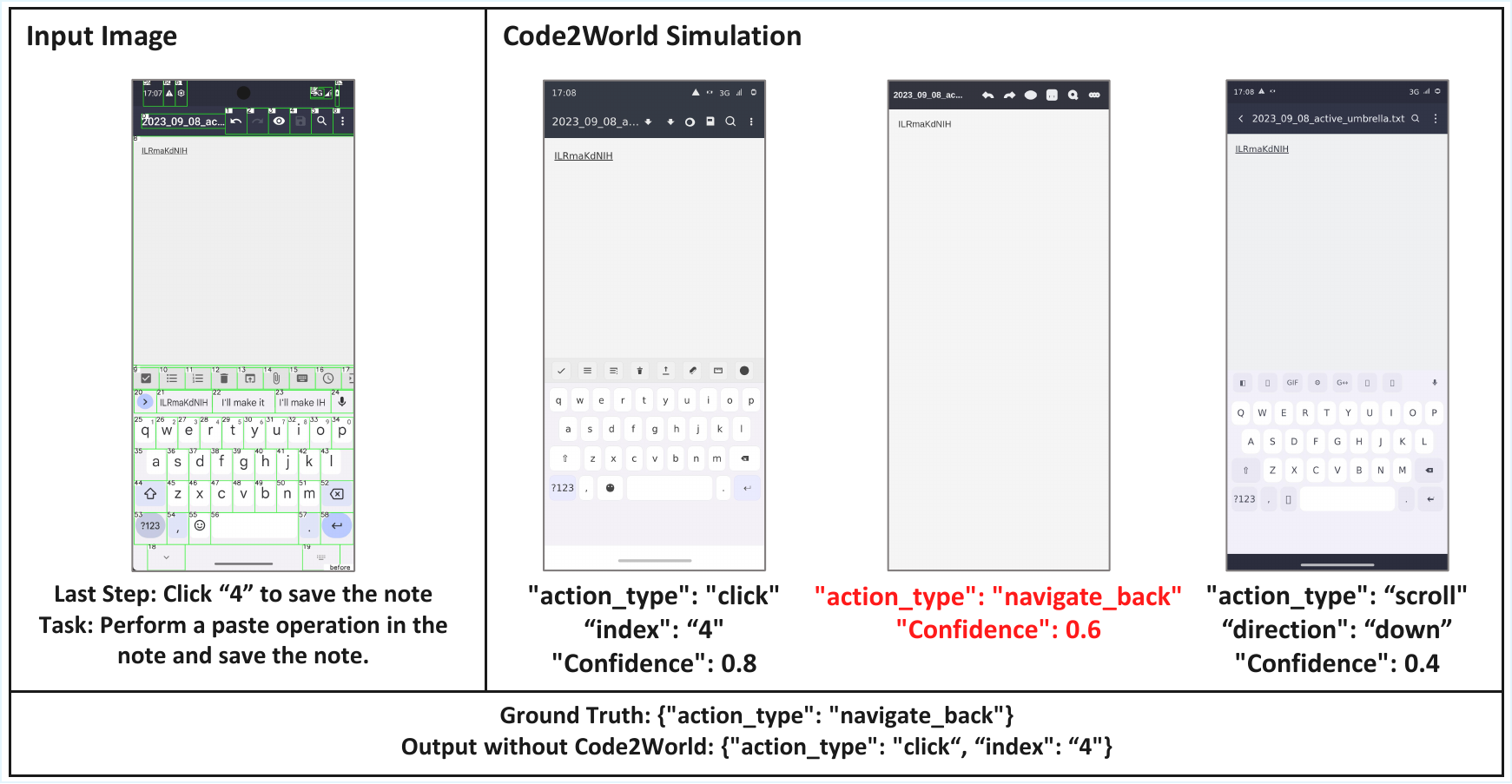}
  \caption{Agent action-selection performance w/ and w/o Code2World on the MarkorCreateNoteFromClipboard task in AndroidWorld. \textcolor{red}{Red} indicates the action ultimately selected by the Code2World pipeline.}
  \label{fig:case1}
  \vspace{-15pt}
\end{figure}

\begin{figure}[t]
  \centering
  \includegraphics[width=\columnwidth]{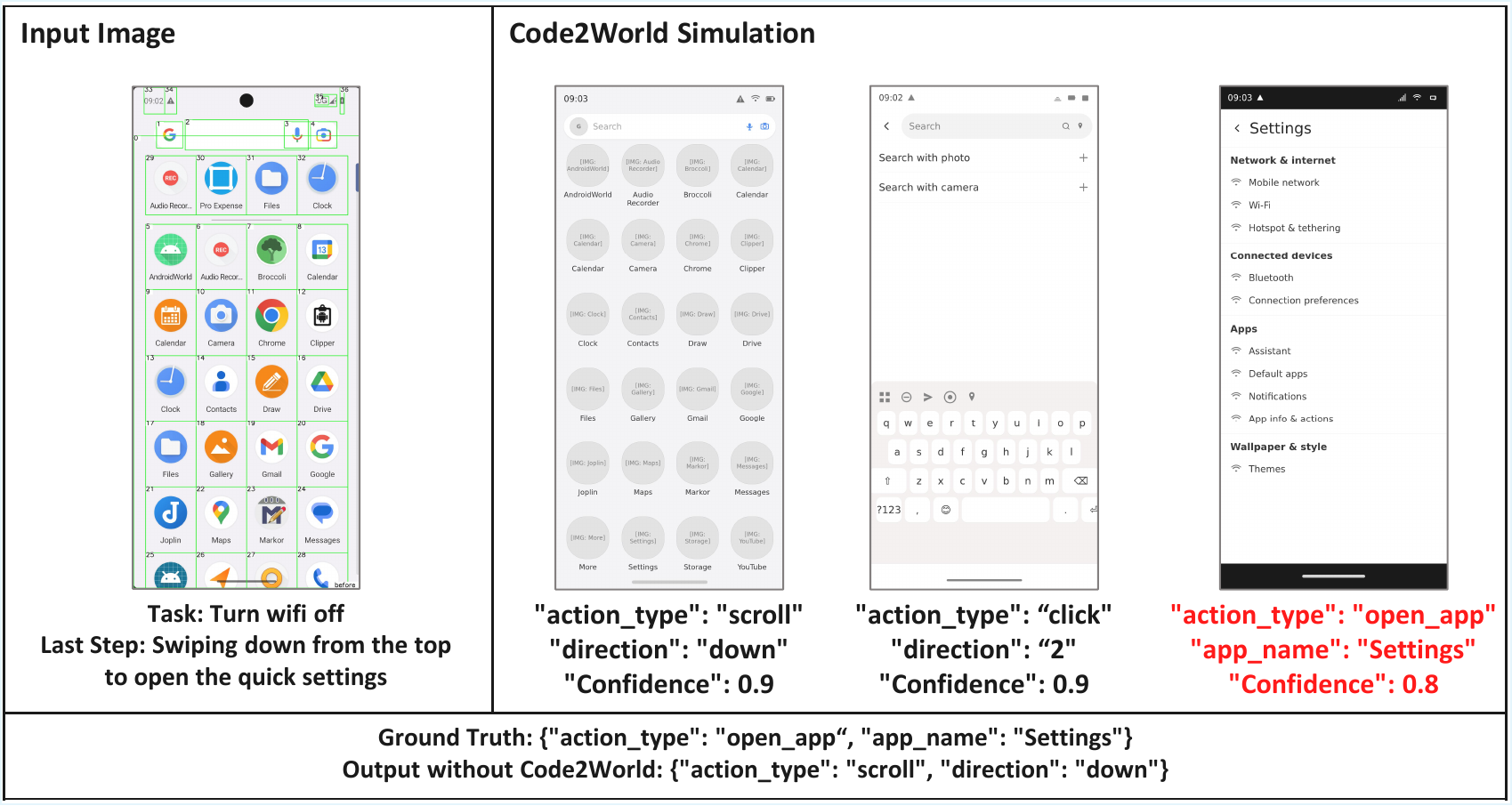}
  \caption{Agent action-selection performance w/ and w/o Code2World on the SystemWifiTurnOffVerify task in AndroidWorld. \textcolor{red}{Red} indicates the action ultimately selected by the Code2World pipeline.}
  \label{fig:case2}
  \vspace{-15pt}
\end{figure}

\begin{figure}[t]
  \centering
  \includegraphics[width=\columnwidth]{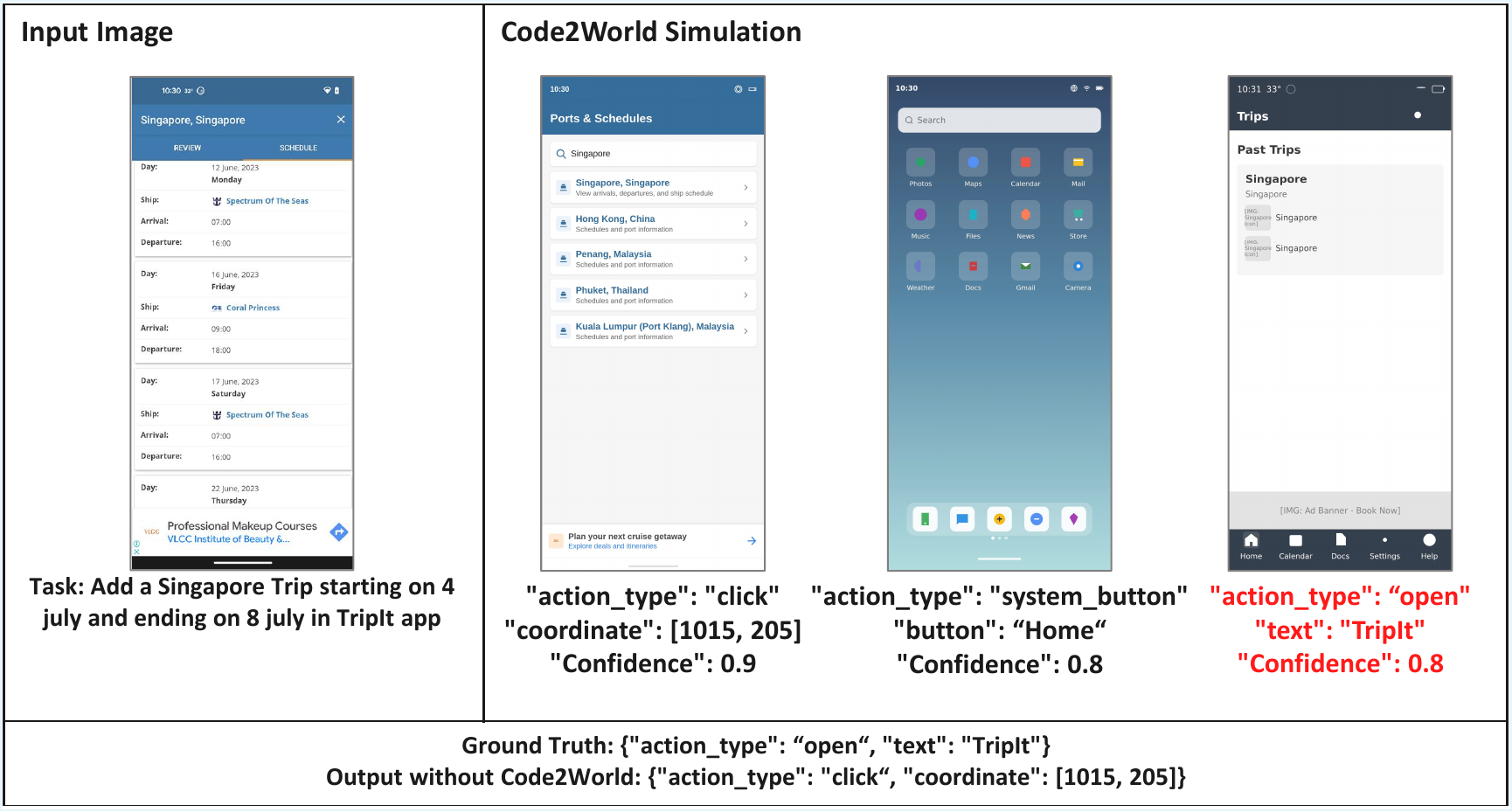}
  \caption{Agent action-selection performance w/ and w/o Code2World at step 0 of Episode 14178 in AndroidControl-High. \textcolor{red}{Red} indicates the action ultimately selected by the Code2World pipeline.}
  \label{fig:case3}
  \vspace{-15pt}
\end{figure}

\begin{figure}[t]
  \centering
  \includegraphics[width=\columnwidth]{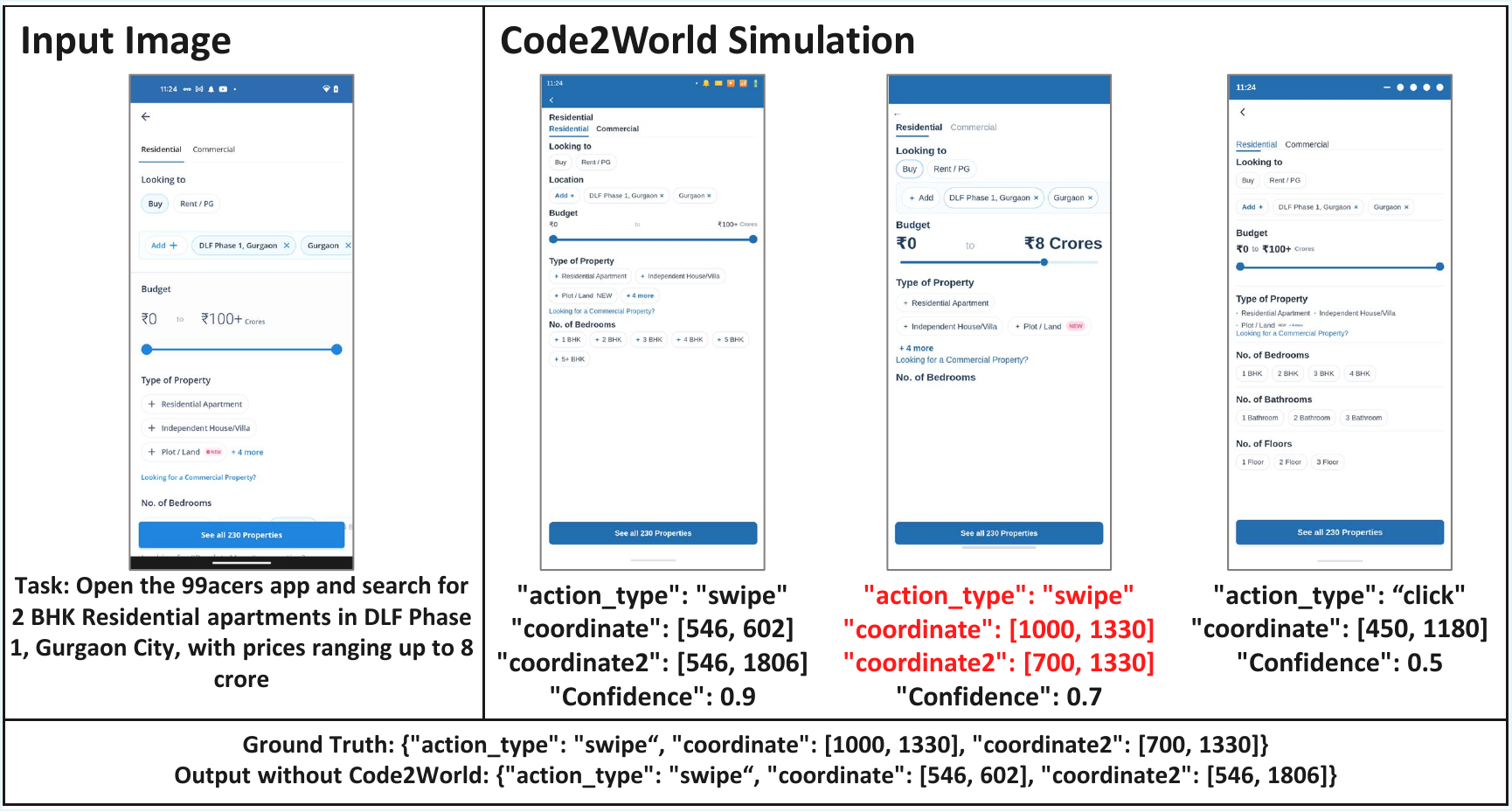}
  \caption{Agent action-selection performance w/ and w/o Code2World at step 7 of Episode 2673 in AndroidControl-High. \textcolor{red}{Red} indicates the action ultimately selected by the Code2World pipeline.}
  \label{fig:case4}
  \vspace{-15pt}
\end{figure}

We show more examples of Code2World enhancing GUI agents in the figure~\ref{fig:case1}-\ref{fig:case4}. As illustrated in Figure~\ref{fig:case1}, the agent had already saved in the last step via the action ‘click 4’, but due to limited visual perception, it failed to detect the change in the save icon, thus intending to click the “Save” button again. Through proactive preview, Code2World enables the model to realize that clicking again will not cause any change, leading it to select a different action, “navigate\_back”, thereby avoiding an unnecessary loop. In Figure~\ref{fig:case2}, on the current application page, to turn off Wi‑Fi, the agent naturally chooses to scroll down to locate the settings app, a correct but inefficient strategy. Under Code2World’s pipeline, the agent is prompted to explore different possible actions, thereby discovering “open\_app”, a more efficient and direct action. Subsequently, Code2World correctly predicts the interface after “open\_app” launches Settings, allowing the agent to more intuitively understand that “open\_app” can reach the Settings page faster, thus completing the task in fewer steps. Similarly, as shown in Figure~\ref{fig:case3}, to locate the TripIt app, the agent explores three different actions. Although all three contribute to task progress, Code2World's predictions indicate that only open\_app directly opens the TripIt application interface. As a result, the agent is able to select the most efficient action "open\_app". In Figure~\ref{fig:case4}, on the detail page, the agent has the instinctive impulse to scroll for more information, but Code2World demonstrates that Action 2 adjusts the price to 8 crore, a result that clearly advances the user’s task and thus prevents a pointless scroll. Moreover, Code2World successfully predicts the future GUIs resulting from Actions 1 and 3, which would scroll to the bottom and click inactive elements without producing any change.

\end{document}